\documentclass[preprint,12pt]{elsarticle}

\makeatletter
\def\ps@pprintTitle{%
 \def\@oddfoot{\footnotesize\fbox{\parbox{\dimexpr\textwidth-2\fboxsep-2\fboxrule\relax}{%
   This article has been accepted for publication in Computer Networks. This is the author's accepted manuscript version. Copyright may transfer without notice.}}}
 \let\@evenhead\@empty
 \let\@oddhead\@empty
 \let\@evenfoot\@oddfoot}
\makeatother

\usepackage{amssymb}
\usepackage{amsmath}

\usepackage{booktabs}
\usepackage{glossaries}

\newacronym{ai}{AI}{Artificial Intellegence}
\newacronym{dl}{DL}{Deep Learning}
\newacronym{ml}{ML}{Machine Learning}
\newacronym{dnn}{DNN}{Deep Neural Network}
\newacronym{iot}{IoT}{Internet of Things}
\newacronym{nas}{NAS}{Neural Architecture Search}
\newacronym{jscc}{JSCC}{Joint Source-Channel Coding}
\newacronym{gan}{GAN}{Generative Adversarial Network}
\newacronym{cgan}{cGAN}{conditional Generative Adversarial Network}

\newacronym{snr}{SNR}{Signal-to-Noise Ratio}
\newacronym{ofdm}{OFDM}{Orthogonal Frequency Division Multiplexing}
\newacronym{mimo}{MIMO}{Multi-Input-Multi-Output}
\newacronym{csi}{CSI}{Channel State Information}
\newacronym{drl}{DRL}{Deep Reinforcement Learning}
\newacronym{awgn}{AWGN}{Additive White Gaussian Noise}

\newacronym{cnn}{CNN}{Convolutional Neural Network}
\newacronym{dag}{DAG}{Directed Acyclic Graph}
\newacronym{semcom}{SemCom}{Semantic Communication}
\newacronym{sl}{SL}{Semantic-Level}
\newacronym{tl}{TL}{Task-Level}
\newacronym{kb}{KB}{Knowledge Base}
\newacronym{lstm}{LSTM}{Long Short-term Memory}
\newacronym{wer}{WER}{Word Error Rate}
\newacronym{ber}{BER}{Bit Error Rate}
\newacronym{bler}{BLER}{Block Error Rate}
\newacronym{ser}{SER}{Symbol Error Rate}
\newacronym{kl}{KL}{Kullback-Leibler}
\newacronym{psnr}{PSNR}{Peak Signal-to-Noise Ratio}
\newacronym{ssim}{SSIM}{Structural Similarity Index Method}
\newacronym{bleu}{BLEU}{Bilingual Evaluation Understudy}
\newacronym{cider}{CIDEr}{Consensus-based Image Description Evaluation}
\newacronym{arq}{ARQ}{Automatic Repeat Request}
\newacronym{phy}{PHY}{Physical Layer}
\newacronym{vqa}{VQA}{Visual Question Answering}
\newacronym{ib}{IB}{Information Bottleneck}
\newacronym{dib}{DIB}{Distributed Information Bottleneck}
\newacronym{tta}{TTA}{Test-Time Adaptation}
\newacronym{ood}{OOD}{Out of Distribution}
\newacronym{sec}{SEC}{Semantic Edge Computing}
\newacronym{vit}{ViT}{Vision Transformer}

\newacronym{isac}{ISAC}{Integrated Sensing and Communication}
\newacronym{oran}{Open-RAN}{Open Radio Access Network}
\newacronym{rf}{RF}{Radio Frequency}
\newacronym{rfmls}{RFMLS}{Radio Frequency Machine Learning System}
\newacronym{llm}{LLM}{Large Language Model}

\newacronym{bfi}{BFI}{Beamforming Feedback Information}
\newacronym{vae}{VAE}{Variational Auto-Encoder}
\newacronym{lpips}{LPIPS}{Learned Perceptual Image Patch Similarity}

\usepackage{lscape}
\usepackage{wrapfig}
\usepackage{xspace}
\usepackage[table]{xcolor}
\usepackage{multirow}
\usepackage{makecell}
\newcommand{\dnn}{\gls{dnn}\xspace}
\newcommand{\dnns}{\glspl{dnn}\xspace}
\newcommand{\rev}[1]{\textcolor{black}{#1}}

\journal{Computer Networks}

\begin{document}

\begin{frontmatter}

\title{Semantic Edge Computing and Semantic Communications in 6G Networks: A Unifying Survey and Research Challenges}

\author[label1]{Milin Zhang\fnref{contrib}}
\ead{zhang.mil@northeastern.edu}
\author[label1]{Mohammad Abdi\fnref{contrib}}
\ead{abdi.mo@northeastern.edu}
\author[label2]{Venkat R. Dasari}
\ead{venkateswara.r.dasari.civ@army.mil}
\author[label1]{Francesco~Restuccia}
\ead{frestuc@northeastern.edu}

\affiliation[label1]{organization={Northeastern University},
            country={United States}}

\affiliation[label2]{organization={DEVCOM Army Research Laboratory},
            country={United States}}

\fntext[contrib]{These authors contributed equally to this research.}


\begin{abstract}
\gls{sec} and \glspl{semcom} have been proposed as viable approaches to achieve real-time edge-enabled intelligence in sixth-generation (6G) wireless networks. On one hand, \gls{semcom} leverages the strength of \glspl{dnn} to encode and communicate the semantic information only, while making it robust to channel distortions by compensating for wireless effects. Ultimately, this leads to an improvement in the communication efficiency. On the other hand, \gls{sec} has leveraged distributed \dnns to divide the computation of a \dnn across different devices based on their computational and networking constraints. Although significant progress has been made in both fields, the literature lacks a systematic view to connect both fields. In this work, we fill the current gap by unifying the \gls{sec} and \gls{semcom} fields. We summarize the research problems in these two fields and provide a comprehensive review of the state of the art with a focus on their technical strengths and challenges. 
\end{abstract}




\date{}

\end{frontmatter}



\glsresetall
\section{Introduction} \label{sec:intro}

For decades, wireless communication has been achieved by encoding information in binaries, introducing redundancy in bitstreams for error correction, and transmitting bits through the unreliable physical layer \citep{shannon1948mathematical}. However, such a bit-level communication paradigm is constrained by the Shannon capacity limit and cannot meet the growing need of data transmission throughput in sixth-generation (6G) wireless networks \citep{luo2022semantic}. For example, for a video-streaming application, sending frames at 120 Hz frame rate with 8K resolution and 8 bits resolution per pixel requires more than 30 Gbps of bandwidth while the IEEE 802.11 ax standard can only provide up to 9.6 Gbps speed.

To this end, the field of \gls{semcom} has investigated data-driven \gls{dl}-based techniques for transmitting signals at the ``semantic'' level. Specifically, \gls{semcom} trains a \dnn-based source and channel encoder and decoder deployed separately at the transmitter and receiver sides, to extract and decode the semantic information that can be directly transmitted over the air \citep{luo2022semantic}. Compared to bit-level communication systems that separately optimize modulation, source and channel coding schemes, \gls{semcom} can achieve more efficient communication through transmission of compressed semantic information and learning-driven joint optimization of modulation and coding schemes.

On the other hand, many emerging \gls{dl}-based technologies, envisioned as potential applications of 6G networks -- i.e., metaverse, smart city, autonomous driving, etc -- require a real-time performance of \dnns which demand significant computational resources. Conventional edge computing that offloads the entire computation to powerful edge servers can result in substantial latency for data transmission. To address this challenge, \gls{sec} is proposed to alleviate the computation and communication overhead of \gls{dl} applications by distributing the workload \citep{matsubara2022split}. The key idea is to split the computation of \dnns across multiple devices based on their computational and networking constraints and transmit latent representations -- i.e., the ``semantic" extracted by the \dnn -- through 6G networks to accelerate the inference. In stark opposition to conventional edge computing where the entire raw data is offloaded to the edge, \gls{sec} reduces the communication overhead by leveraging a bottleneck structure at the splitting point of \dnn to compress the latent representations, hence achieving lower latency \citep{eshratifar2019bottlenet,shao2020bottlenet++,matsubara2022bottlefit}.

While these two fields are derived independently for different purposes, they share notable similarities in their goals and methodologies, which are further discussed in Section~\ref{sec:overview}. For instance, both \gls{sec} and \gls{semcom} leverage \dnns distributed on separate devices for extracting and compressing semantic information that needs to be transmitted in wireless networks. \gls{sec} primarily focuses on optimizing computation, while \gls{semcom} emphasizes the resilience of semantics in wireless channels. Merging techniques from both fields is promising to achieve more efficient and reliable \gls{dl} applications in 6G networks.

\textbf{However, the key issue is that despite the striking similarities in goals between the two, the fields of \gls{sec} and \gls{semcom} have largely ignored each other}. For example, although there is some recent work in \gls{sec} which considered adaptive latent feature compression to varying channel conditions \citep{assine2023slimmable}, it leveraged conventional network protocols such as Wi-Fi to transfer latent representations and neglected advanced coding scheme in \gls{semcom}. Similarly, while prior work has leveraged \gls{ib} theory to compress the encoded semantics \citep{xie2023robust,pezone2022goal}, most work in \gls{semcom} used \dnns with custom-tailored fixed architectures to build communication systems without considering recent feature compression techniques in \gls{sec}. \textbf{To fill the existing gap, this paper provides a unified view of  \gls{sec} and \gls{semcom} to demystify their similarities and summarize the latest breakthroughs in both areas. }Our work is orthogonal to existing literature reviews \citep{matsubara2022split,luo2022semantic} since the unified overview helps both research fields to discover the similarity and adopt a new perspective from the other community. \smallskip

\noindent\rev{\textbf{Scope and Contributions.} In this survey, we focus on the technologies that aim to achieve edge intelligence through collaborative execution of \glspl{dnn} during inference time. Our scope deliberately excludes federated learning and split learning which focus on distributing \gls{dnn} computation during the training phase.} 

\rev{Several existing studies address related topics such as split computing, collaborative inference, and \gls{dnn} partitioning \cite{matsubara2022split,ren2023survey,wang2024end,zhang2025survey} for distributing computation during inference to accelerate \gls{dl} applications. For example, \citet{matsubara2022split} summarized the split computing methods to accelerate \gls{dnn} applications in edge computing. \citet{ren2023survey} focused on optimizing the \gls{dnn} computation allocation in the device-edge-cloud collaborative environment. \citet{wang2024end} reviewed the security and privacy challenges of this distributed \gls{dnn} system during the communication across devices. \citet{zhang2025survey} summarized the training, inference and knowledge transfer technologies in mobile-edge-cloud computing.}

\rev{Despite overlapping scopes, these topics typically emphasize different aspects of edge computing and lack a comprehensive perspective. For example, split computing primarily focuses on optimizing \gls{dnn} architecture and training methods to minimize computation on resource-limited devices while preserving task performance. In contrast, cooperative inference or \gls{dnn} partitioning concentrates on resource allocation within multi-device collaborative systems for deploying pre-trained, off-the-shelf \glspl{dnn}. Compared to other work that solely focuses on split computing \cite{matsubara2022split} or \gls{dnn} partitioning \cite{ren2023survey,wang2024end}, our survey aims to provide a more holistic view of these interconnected approaches. As such, we introduce the term ``\acrfull{sec}" to encompass these concepts, acknowledging their shared foundation of transmitting latent representations (i.e., ``semantics") extracted by \glspl{dnn} for task execution. }

\rev{In addition, \gls{semcom} is recognized as a promising approach to enable \gls{dl} deployment at the edge. Several surveys explore different aspects of \gls{semcom}. For instance, \citet{yang2022semantic} reviewed the theory development and proposed a categorization of \gls{semcom} technologies. \citet{luo2022semantic} explained the \gls{semcom} concept and studied the \gls{dl} progress in this field. \citet{guo2024survey} highlighted the security and privacy challenges of \gls{semcom} in a multi-agent setting while \citet{getu2025semantic} reviewed the current progress with an emphasis on 6G applications. However, existing surveys overlook related research in \gls{sec}, which could potentially enhance \gls{semcom} through the integration of sophisticated feature extraction techniques and system optimization strategies. Our work separates itself by providing a new framework that connects \gls{semcom} to \gls{sec} with a detailed discussion on the methodology and challenges. Table~\ref{tab:comparison} summarizes our main contribution and distinction from other relevant surveys.} \smallskip

\noindent\textbf{Paper Organization.}~Section~\ref{sec:overview} introduces the background and an overview to discuss the relation between \gls{sec} and \gls{semcom}. Section~\ref{sec:sec} and Section~\ref{sec:semcomm} summarize the recent progress in \gls{sec} and \gls{semcom}, respectively. We conclude our paper by discussing the limitations and future directions in Section~\ref{sec:discuss} and providing a summary in Section~\ref{sec:conclude}. 

\begin{landscape}
\begin{table}
\caption{A Comparison between Relevant Survey and Our Work}
    \centering
    \resizebox{\linewidth}{!}
    {
    \begin{tabular}{c|c|c|c}
    \toprule
    
    Category &  Year & Key focus and contributions & Our contributions and distinctions \\
       \hline\hline
      \multirow{4}{1.5cm}[-5ex]{SemCom} & 2022 \cite{yang2022semantic} & \makecell{It provides a roadmap of SemCom theory and \\categorization of SemCom technologies with \\a discussion on their benefit and limitations.} & \multirow{4}{*}{
       \makecell{
       $\bullet$ We categorize SemCom, and propose \\a new framework to connect SemCom \\and SEC. \\
       $\bullet$ We propose designing methods to this \\framework, addressing its advantages \\and challenges. \\
       $\bullet$ We provide a comprehensive literature \\review of recent progress in SemCom \\technologies. \\
       }
       } \\
       
       \cline{2-3}
       & 2022 \cite{luo2022semantic} & \makecell{It explains the SemCom concept and reviewed \\ recent deep learning in this emerging field.} & \\
       
       \cline{2-3}
       & 2025 \cite{getu2025semantic} & \makecell{It summarizes the SemCom framework and \\current research with an emphasis on \\6G applications.} & \\
       
       \cline{2-3}
       & 2024 \cite{guo2024survey} & \makecell{It introduces a three-layer framework \\for multi-agent SemCom and explores \\security and privacy issues.} & \\
       
       \hline\hline
       
       \multirow{4}{1cm}[-8ex]{SEC} & 2022 \cite{matsubara2022split} & \makecell{It reviews tasks, technologies and models \\of split computing and early exiting \\in device-edge computing} & 
       \multirow{4}{*}{
       \makecell{
       $\bullet$ We summarize SEC, including split \\computing, cooperative inference, and \\DNN partitioning. \\
       $\bullet$ We categorize SEC based on their \\technical focus, revealing the \\connection to SemCom. \\
       $\bullet$  We provide a comprehensive review \\of the methodologies, technical \\challenges of SEC. \\
       $\bullet$  We discuss limitations of current \\SemCom and SEC and potential \\future research directions. \\
       }
       }
       \\
       
       \cline{2-3}
        & 2023 \cite{ren2023survey} & \makecell{It summarizes device-device, device-edge, \\edge-cloud collaborative inference with \\a focus on optimization technologies} & \\
       
       \cline{2-3}
       & 2024 \cite{wang2024end} & \makecell{It reviews the security and privacy \\challenges and  technologies in edge-\\cloud DNN partitioning} & \\
       
       \cline{2-3}
       & 2025 \cite{zhang2025survey} & \makecell{It focuses on model training, inference \\and updating in mobile-edge-cloud \\computing, with an emphasis on \\compression, partitioning and knowledge \\transfer technologies} & \\
    \bottomrule
    \end{tabular}
    }
    \label{tab:comparison}
\end{table}
\end{landscape}

\section{Background of Semantic Edge Computing and Semantic Communication} \label{sec:overview}

\subsection{\rev{Edge Intelligence in 6G Network}} \label{sec:background}

\noindent\rev{\textbf{6G Network for Intelligence.}~The large network capacity and massive connectivity of 6G networks is expected to facilitate many \gls{dl}-driven applications such as \gls{isac} \cite{haque2023simwisense,haque2025beamsense,haque2024bfa}, metaverse \cite{yu2023semantic,luong2023incentive} and autonomous driving \cite{huang2025task,liu2025cross,ma2024survey}. \gls{isac} can achieve object tracking, localization, and activity recognition using the \gls{csi} \cite{haque2023simwisense} or \gls{bfi} \cite{haque2024bfa} of the signal which carries data traffic. Compared to conventional sensing methods that need independent infrastructure, \gls{isac} leverages the existing communication signals and spectrum to perform sensing, creating a unified system where communication and sensing capabilities are intrinsically merged within the same hardware and spectrum resources. In \gls{isac}, \glspl{dnn} are essential for processing and interpreting the complex patterns within \gls{csi} or \gls{bfi} to distinguish between different objects, movements, and environmental changes. The large bandwidth and high-volume data transmission capabilities of 6G networks can enhance the available object information from carrier signals, enabling significantly improved sensing resolution.}

\rev{In addition, metaverse relies heavily on \gls{dl} algorithms for a continuous 3D environment rendering, human pose estimation and spatial tracking while autonomous driving leverages \glspl{dnn} for object detection, dynamic path planning, and decision-making based on massive sensor data streams. All these applications require a massive data transmission and processing. As such, the network capacity of 6G networks can enable seamless real-time interactions in immersive metaverse environments while providing the instantaneous data processing and communication required for safe, coordinated autonomous vehicle networks.} \smallskip

\noindent\rev{\textbf{Intelligence for 6G Network.}~On the other hand, 6G is conceived as an intelligent network where \gls{dl} is integrated as a foundational layer, enabling adaptive, self-optimized and massive communication. For instance, \gls{rfmls} is a new signal processing paradigm that applies modern machine learning techniques and \glspl{dnn} to the \gls{rf} spectrum domain. It leverages the strength of \gls{dnn} to process \gls{rf} waveforms, identify different signals and detect anomalies \cite{mittal2024sub,uvaydov2024stitching,zhang2024hyperadv}. As such, it enables efficient spectrum management where the transmitter can dynamically adjust parameters such as power levels, modulation schemes, and frequency allocations based on real-time spectrum conditions and varying traffic demands, significantly improving spectral efficiency and network performance.}

\rev{In addition, \gls{oran} \cite{puligheddu2023sem,baldesi2022charm,maxenti2024scalo}, which aims to transform traditional RAN into open, standardized, and interoperable elements, is expected to be the foundation of 6G network. Compared to conventional RANs that are highly integrated and vendor-specific, \gls{oran} creates independent, modularized interfaces for flexible connections to different network functions. It uses RAN Intelligent Controllers (RICs) to seamlessly integrate various \gls{dl} solutions for autonomous network resource management and performance optimization.} \smallskip

\noindent\rev{\textbf{Edge Intelligence.} However, achieving intelligence in 6G networks faces critical computation and communication challenges. Research has demonstrated that \glspl{dnn} harvest performance from scaling-up model size \cite{kaplan2020scaling} while many 6G technologies need to be deployed on resource-constrained devices. A possible solution is to create lightweight, hardware-efficient \gls{dnn} architectures. However, these compressed models typically exhibit substantial performance degradation. For example, MobileNetV2 suffers more than 3\% accuracy loss on ImageNet classification compared to ResNet50 \cite{sandler2018mobilenetv2}. While offloading computation to powerful cloud servers represents a viable solution to overcome local resource constraints, this approach introduces massive communication overhead. For instance, metaverse requires the \gls{dl} algorithm to continuously process the high quality video streams to create an immersive AR/VR environment. Offloading large volumes of data to edge/cloud can result in significant latency. To this end, \gls{sec} and \gls{semcom} are proposed to bring intelligence to the edge by distributing computation across multiple devices and minimize the inter-device communication overhead.}

\subsection{Semantic Edge Computing} \label{sec:overview_sec}

\begin{figure}
    \centering
    \includegraphics[width=0.9\linewidth]{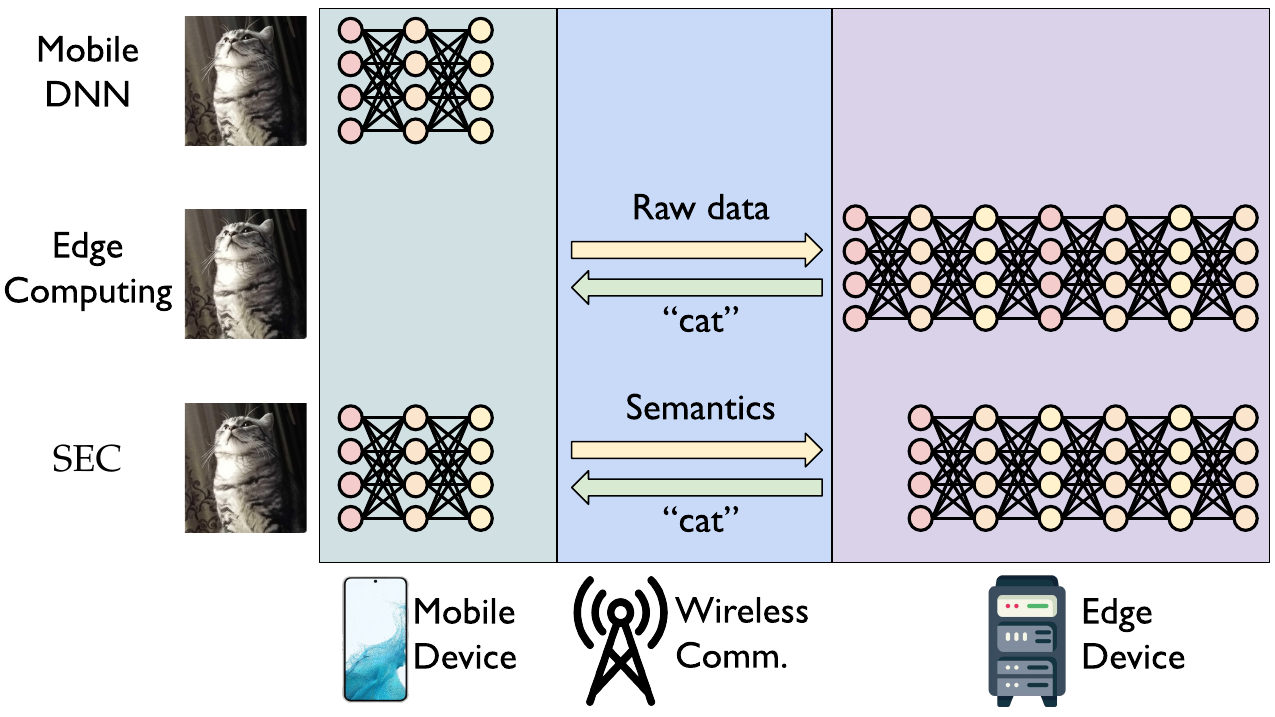}
    \caption{Comparison of mobile DNN, edge computing and semantic edge computing.}
    \label{fig:sec}
\end{figure}

\rev{To address the computation and communication challenge in deploying \dnns at edge, prior art proposed ``\dnn partitioning" \cite{feltin2023dnn} and ``split computing" \cite{matsubara2022bottlefit} to accelerate \dnn inference by leveraging computational resources across devices in a cooperative manner. The key idea of \dnn partitioning is to divide \dnn layers across multiple devices to overcome the computation burden that no single resource-constrained device could handle. Compared to conventional edge offloading that transmits raw input, the collaborative inference requires the system to send latent representations -- i.e., the ``semantic" -- generated at the \dnn splitting point. \dnn partitioning concentrates on complex optimization with various conditions such as network bandwidth, device computational capabilities, energy consumption, and latency requirements, with the goal of minimizing end-to-end inference time and ensuring efficient resource utilization.}

\rev{However, simply splitting \dnns may result in excessive communication burden as the latent representation can have large data size without compression. For example, the intermediate representation in early layers of ResNet can go up to 5x larger than the original input data \cite{shao2020communication}. As such, research in split computing focuses on \dnn architecture design and advanced training mechanism to compress the latent representation at the splitting point while maintaining sufficient semantic information for task performance.}

\rev{Despite their high relevance, existing literature in split computing and \dnn partitioning focus on narrow perspectives. In this survey, we use ``\acrfull{sec}" to unify these overlapping topics and summarize the state of the art in each field. We adopt the term ``semantic" to distinguish this approach from traditional edge computing paradigms that offload entire computational tasks to remote servers and transmit raw input data for processing. In contrast, SEC selectively offloads specific computational components while transmitting semantically meaningful features extracted by intermediate layers, rather than unprocessed data. Formally, \gls{sec} refers to a designing pattern that a \dnn $\mathcal{M}$ can be modeled as a Markov chain $\mathcal{M}: X\mapsto L_{1}\mapsto L_{2}\mapsto \cdots \mapsto L_{i}\mapsto\cdots\mapsto Y$. $X$, $Y$ and $L_{i}$ are input data, output data, and the latent representation of $i$-th hidden layer. Without loss of generality we assume the Markov chain is split into a head model $\mathcal{H}: X\mapsto\cdots\mapsto L_{i}$ and a tail model $\mathcal{T}: L_{i}\mapsto\cdots\mapsto Y$, deployed separately on the mobile and edge device. }

\rev{As summarized in Figure \ref{fig:sec}, compared to mobile \dnns, \gls{sec} can leverage the computational power on edge devices to achieve better \gls{dl} performance. \gls{sec} also achieves less latency compared to conventional edge computing by transmitting semantics that have a smaller size than raw inputs \citep{matsubara2022split}. \gls{sec} aims to achieve the following goals: (1) optimize the computation load across mobile and edge devices; and (2) compress the semantic information for inference that needs to be transmitted to reduce the communication latency.}

\subsection{Semantic Communication} \label{sec:bg-semcom}

\begin{figure}[t]
    \centering
    \includegraphics[width=0.95\linewidth]{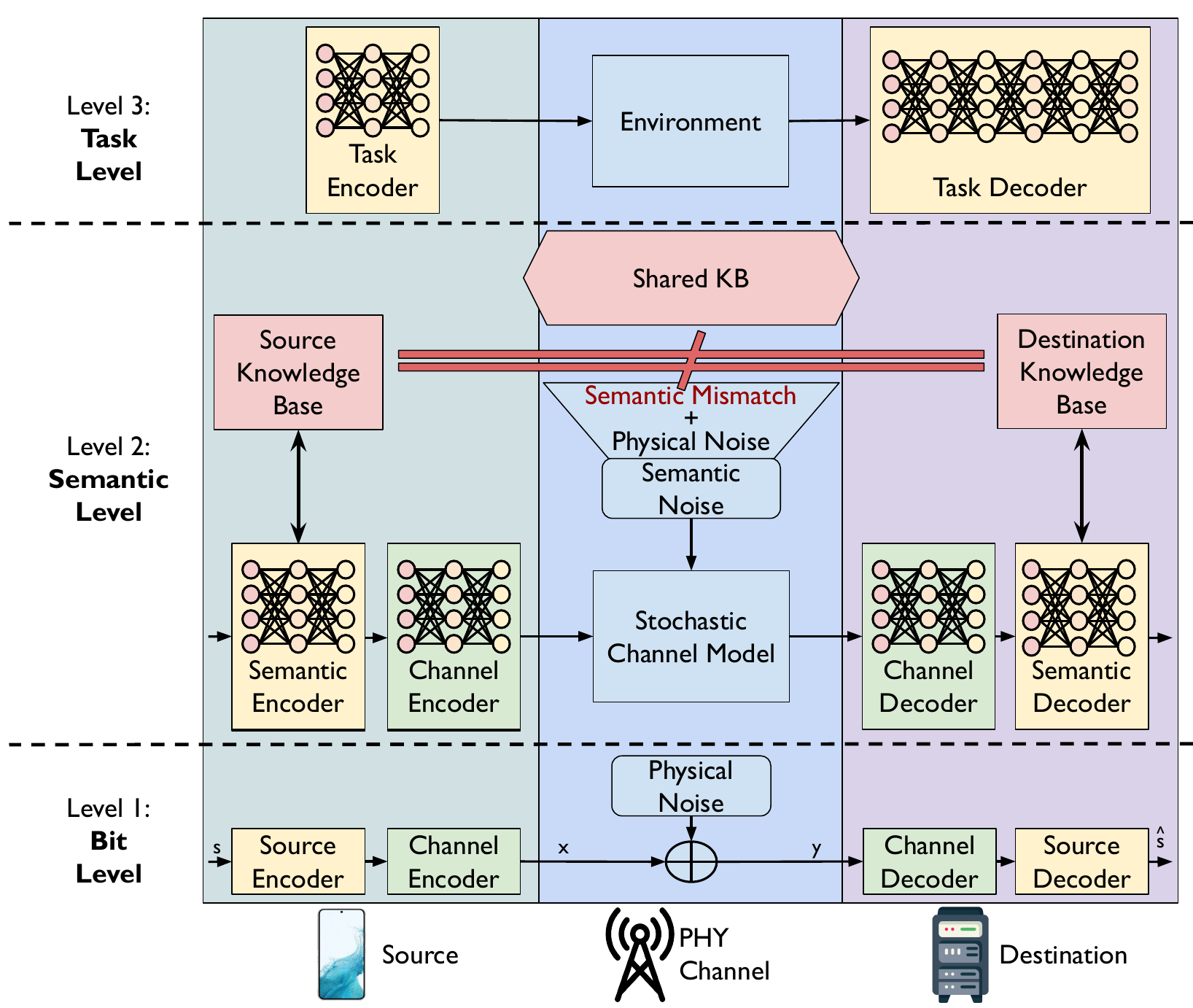}
    \caption{The three levels of communication systems.}
    \label{fig:semcom}
\end{figure}

\Gls{semcom} is a paradigm shift in building communication systems from traditional architectures committed to the accurate transmission of the message in its bit-level representation regardless of its meaning to modern ones which focus on the content/relevance and transmit only the semantics of the message. In general, communication systems can operate at three different levels as illustrated in Figure~\ref{fig:semcom}.

The first level -- corresponding to the classical communication systems and called the bit level -- is designed to reliably transfer the symbolic representation of data from the transmitter to the receiver. There is vast research investigating this type of communication which usually uses conventional independent and separately-optimized blocks to develop the system \cite{yehongyan2020deep, he2020model, erpek2018learning}. In classical communication systems, a transmitter, which aims to send a message $s$, encodes it into $x$ using separate source and channel coding blocks and transmits it via a channel. The intended receiver receives a distorted signal $y$, and tries to accurately decode the message $\hat{s}$ in terms of the bit-level metrics. Assuming the channel matrix to be $H$, the received signal can be derived as $y = Hx + \epsilon$, where $\epsilon$ denotes the channel noise.

The second type of systems -- called the semantic-level systems -- are focused on extracting the semantics of data and often use \dnns both at the transmitter and receiver which are jointly trained in an end-to-end manner incorporating the channel model \cite{farsad2018deep, bourtsoulatze2019deep, xie2021deep, lu2021reinforcement}. Such systems can break the Shannon's limit by transmitting only the semantic meaning behind the message instead of raw bits. As such, as long as the receiver can understand and infer the expected important information correctly, the loss of semantically irrelevant and redundant content can be tolerated. Note that while semantic and channel coding blocks are shown separately in Figure~\ref{fig:semcom}, both functionalities can be merged together which is also referred to as \gls{jscc} in the literature. More specifically, the semantic encoder extracts semantic features using the integrated \gls{kb} from the input $s$ and filters out the redundant information. Then, the channel encoder generates the channel symbols $x$, which are sent through the channel. During transmission, the symbols $x$ are distorted according to the semantic noise accounting for the physical noise, semantic mismatch between the source and destination \glspl{kb}, ambiguity and interpretation errors. Semantic noise extends the concept of physical noise in classical communication to any noise causing semantic distortion. At the receiver, the channel decoder estimates the semantically corrupted features using the received symbols $y$, and the semantic decoder recovers the semantically similar message $\hat{s}$ based on the destination \gls{kb} by maximizing the semantic similarity metrics \cite{zhou2021semantic}. Therefore, both the transmitter and receiver require to share and continuously update a \gls{kb} using the knowledge from the environment.

The third level -- called the task level -- aims to effectively utilize the semantic information to execute a processing task initiated by the transmitter and finalized at the receiver side \cite{xie2021task, kountouris2021semantics, yuan2021graphcomm}. As such, the goal in this level is to perform a task such as an \gls{ai} task on top of the extracted semantics, and thus the developed systems are evaluated by effectiveness metrics such as inference task performance \cite{yuan2021graphcomm, kountouris2021semantics}, energy efficiency \cite{chen2023neuromorphic}, age of information \cite{yates2021age}, and value of information \cite{ayan2019age}. The \gls{sl}-\gls{semcom} and \gls{tl}-\gls{semcom} are both considered as semantic communication systems.

Most of the existing work views the entire \gls{semcom} system as a \gls{dl}-based reconstruction system, and use an encoder-decoder architecture for the \dnn with the bottleneck representing the encoded data to be transmitted over the channel. Also, a stochastic model representing the channel effects is incorporated between the encoder and decoder \dnn, and the communication system is trained in an end-to-end manner to maximize the similarity metrics between the encoder input and the decoder output. As a result, the \dnns learn the channel distortion patterns and make the encoded symbols robust to wireless impairments. In general, \gls{semcom} focuses on: (1) how to effectively extract the semantic information and encode such semantics into channel symbols; and (2) how to make the semantics robust to the noisy wireless channel.

\subsection{Unified Overview}\label{sec:unify}

\begin{figure}[t]
    \centering
    \includegraphics[width=0.95\linewidth]{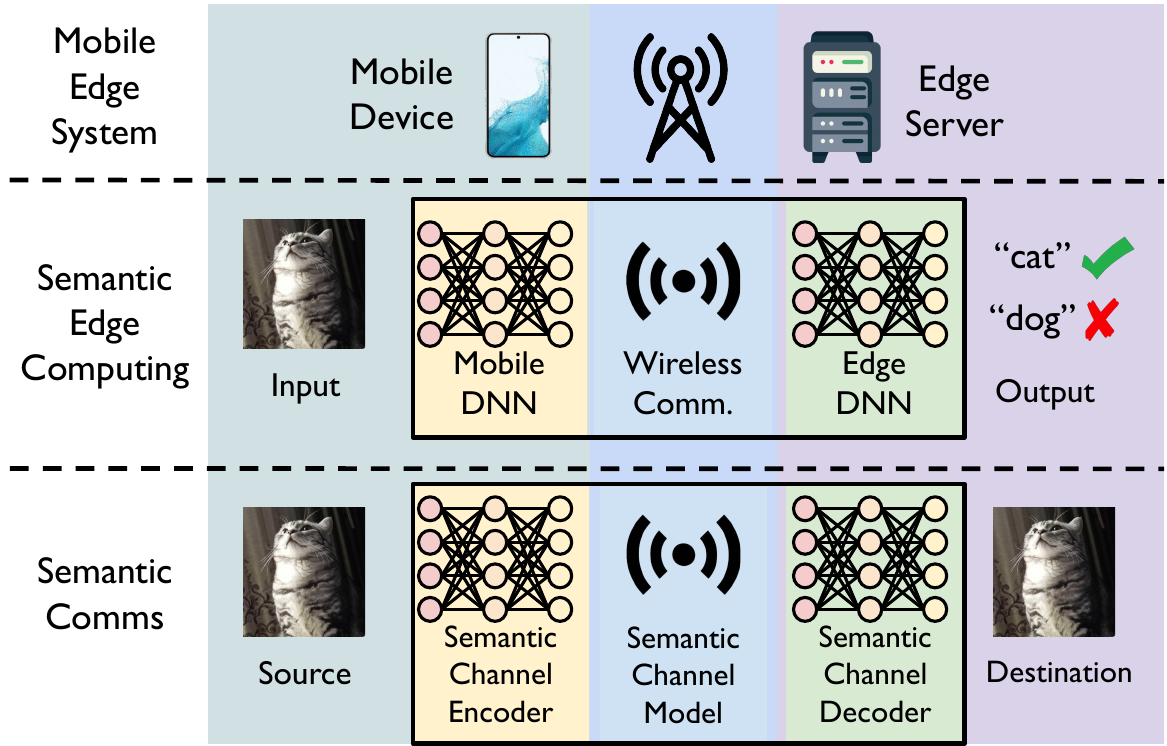}
    \caption{An overview of Semantic Edge Computing (SEC) and Semantic Communication (SemCom).}
    \label{fig:overview}
\end{figure}

Although \gls{sec} and \gls{semcom} are derived independently, they share significant similarities in achieving fast and reliable edge intelligence. Figure~\ref{fig:overview} overviews \gls{sec} and \gls{semcom}. \gls{semcom} consists of a \dnn on the mobile device (transmitter) for semantic and channel coding and a \dnn on the edge device (receiver) for decoding. The encoder and decoder \dnns are trained in an end-to-end fashion to maximize a semantic similarity metric between the source and destination messages. Meanwhile, the mobile \dnn and edge \dnn in \gls{sec} -- which are used to extract the necessary information from the input and decode the information for inference -- are equivalent to the encoder-decoder framework in \gls{semcom}. Indeed, \gls{tl}-\gls{semcom} could be considered as a type of \gls{sec} that aims to achieve task-oriented communication by leveraging distributed \dnns.

\rev{However, existing research in each field often overlooks the development in the other domain,  resulting in suboptimal approaches for achieving intelligence in 6G networks. \gls{semcom} research typically disregards resource constraints in mobile devices and use a large encoder structure to perform more effective semantic extraction. For instance, \cite{shao2021learning} utilized a large ResNet backbone for semantic feature extraction on mobile devices while deploying only two linear layers for decoding at the edge server for tiny ImageNet classification tasks. While these approaches improve the inter-device communication efficiency, they significantly increase the computation overhead on the mobile device and do not make fully use of the computation power at the edge server. On the other hand, \gls{sec} methods such as \cite{eshratifar2019jointdnn,liang2023dnn} only leverage existing wireless communication technologies to transmit latent representations. These methods require full-stack processing to achieve resilience in wireless channels which can introduce additional communication latency.} 

\rev{Therefore, we provide a comprehensive review of both domains and propose to incorporate \gls{sec} with \gls{semcom} across three dimensions: system level, application level, and physical level. From a system optimization perspective, the collaborative operation of \glspl{dnn} across multiple devices involves heterogeneous computation capacity and network conditions. Therefore, this dimension focuses on optimizing computational and communication load on each device to achieve optimal inference latency and energy consumption. Many optimization methods in \gls{sec} such as linear programming \cite{gao2019deep} and \gls{drl} \cite{li2024optimal} can be applied to optimize the utilization of resources. In addition, system level optimization also encompasses the task scheduling and resource sharing to optimize the overall throughput which is missing in multi-view \cite{shao2022task} or multi-user \cite{xie2022task} \gls{semcom}. }

\rev{Application level optimization encompasses the concept of semantic extraction and compression in \gls{sec} and \gls{semcom}, with its emphasis on the performance of \dnn applications. The objective is to train \dnns that can effectively compress the semantic information to be sent while preserving task-oriented performance. Model compression methods, including knowledge distillation \cite{hinton2015distilling}, \gls{nas} \cite{ren2021comprehensive} and pruning \cite{cheng2024survey} can be applied to compress the data size. In addition, application level optimization can be enhanced through carefully designed loss functions that balance compression efficiency with task performance \cite{matsubara2022supervised} or through multi-stage advanced training methodologies \cite{matsubara2022bottlefit}. }

\rev{Physical level optimization involves a cross-layer design to overcome the physical layer distortions. Compared to current research in \gls{sec} that leverages existing wireless protocols, \gls{semcom} methods that directly encode semantic information for wireless transmission can substantially decrease end-to-end latency by bringing \glspl{dnn} to the physical layer \cite{abdi2025phydnns}. Such an approach requires less processing time to execute the full-stack protocol and enhance the reliability of communication by directly incorporating distortions during \dnn training. In addition, physical level optimization involves adaptive transmission techniques \cite{huang2022real,hossain2023flexible} that selectively transmit semantic information based on content importance and channel conditions can substantially enhance energy efficiency by reducing the transmission of non-critical data over unreliable channels. }

\rev{Note that these three dimensions should not be optimized independently but require joint optimization to achieve optimal performance, representing a significant challenge in current state-of-the-art approaches. We hope our discussion can facilitate future investigations in integrating these design methodologies.}

\section{Recent Progress in Semantic Edge Computing} \label{sec:sec}

\subsection{Taxonomy of Semantic Edge Computing}

\begin{figure}
    \centering
    \includegraphics[width=\linewidth]{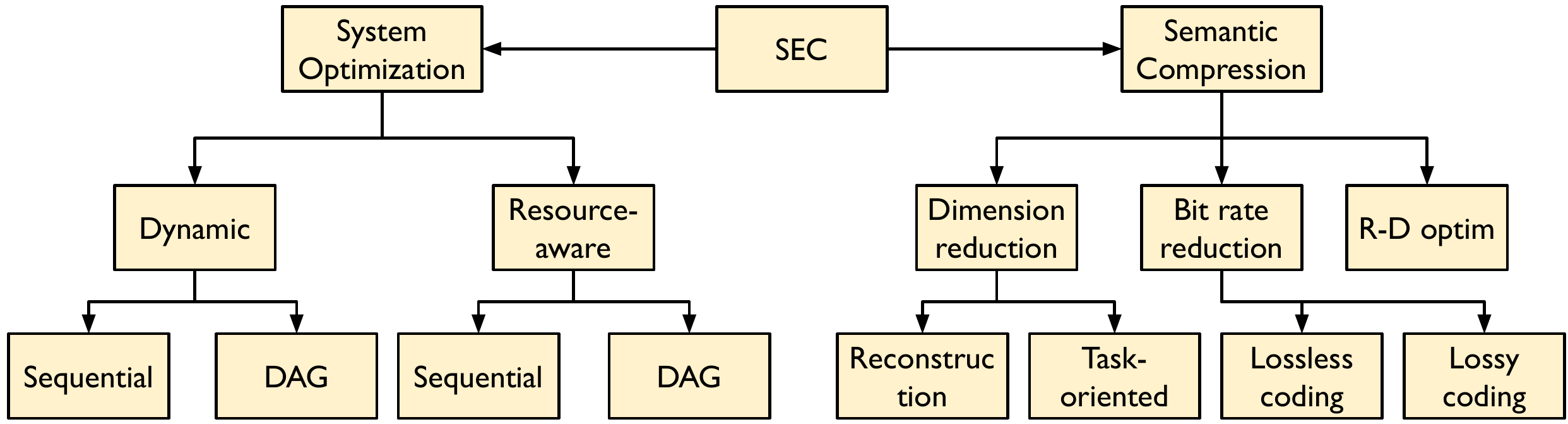}
    \caption{Taxonomy of Recent Progress in Semantic Edge Computing}
    \label{fig:t-sec} 
\end{figure}

As discussed in Section~\ref{sec:overview_sec}, \gls{sec} primarily focuses on two problems: (1) how to optimize the computation of the mobile-edge system based on their resource constraints; and (2) how to extract meaningful semantics while compressing task-irrelevant information to reduce the transmission overhead. Based on these two directions, we categorize recent progress in \gls{sec} into system optimization and semantic compression. In system optimization, literature can be further categorized as static resource-aware optimization and dynamic model partitioning. In each category, \dnns can be distributed based on their sequential topology or \gls{dag} topology. Based on the compression techniques investigated in the literature, semantic compression can be further classified as dimension reduction, bit rate reduction and learning-driven rate-distortion optimization. Dimension reduction aims to introduce bottleneck layers at the splitting point of \dnns to compress the dimension of semantics, which can be further classified as reconstruction-based and task-oriented compression. Bit rate reduction leverages either lossless or lossy coding techniques to reduce the data size that needs to be transmitted. The taxonomy of \gls{sec} is depicted in Figure~\ref{fig:t-sec}. 

\subsubsection{System Optimization}

To achieve minimum end-to-end latency \cite{gao2019deep}, energy consumption \cite{kang2017neurosurgeon} or to meet other Quality-of-Service constraints \cite{zawish2022towards}, the mobile-edge system needs to be jointly optimized with consideration of computing and networking resources. A general working flow is to first estimate each \dnn module cost \cite{kang2017neurosurgeon} or profile the cost of each module on real devices \cite{eshratifar2019jointdnn}. Based on the \dnn module cost, an optimization problem is then formulated and solved with different approaches. For example, the end-to-end latency $T$ can be measured by $T_{c}^{m}(L_i)$, $T_{c}^{e}(L_i)$ and $T_{n}(L_i)$ denoting computing latency at mobile device, edge device, and networking latency respectively where $L_i$ is the splitting point at the $i$-th \dnn layer. The computing resource that is required to execute the head and tail model on mobile and edge devices are denoted as $R_c^{m}(L_i)$, $R_{c}^{e}(L_i)$, which is constrained by the maximum available computing resource on each devices $\Tilde{R}_{c}^{m}$ and $\Tilde{R}_{c}^{e}$. Therefore, a \dnn partitioning optimization aiming to minimize latency can be formulated as the following problem:
    \begin{subequations} \label{eqn:opt}
\begin{align}
    \min_{L_i} \quad & T_{c}^{m}(L_i) + T_{c}^{e}(L_i) + T_{n}(L_i) \\
    s.t. \quad & R_c^{m}(L_i) \leq \Tilde{R}_{c}^{m} \\
    & R_c^{e}(L_i) \leq \Tilde{R}_{c}^{e}
\end{align}     
\end{subequations}
The system optimization problem such as Eqn.~\ref{eqn:opt} is often NP-hard, hence requiring advanced approaches to find the optimal solution. Based on the optimization approach, literature can be categorized as heuristic search \cite{wang2021accelerate}, solver-based \cite{gao2019deep}, relaxation-based \cite{eshratifar2019jointdnn} as well as \gls{drl}-based approaches \cite{li2024optimal}. Based on the partitioning graph, it can also be classified as sequential-based \cite{gao2021task} and \gls{dag}-based \cite{li2021throughput} partitioning. 

In Section~\ref{sec:sys_opt}, we categorize the recent progress on system optimization into resource-aware partitioning and dynamic model partitioning. The resource-aware partitioning aims to achieve the best resource usage (e.g., computing resource and networking bandwidth) by finding the optimal splitting point with different algorithms \cite{feltin2023dnn}. Additionally, as the resource constraints and quality of service requirement may change over time, dynamic model partitioning is proposed to adaptively split \dnn under different conditions \cite{bakhtiarnia2023dynamic}.

\subsubsection{Semantic Compression} 

The transmission overhead in \gls{sec} can be reduced by compressing the dimension and bit-width of semantic representations. For example, a general approach to add an auto-encoder to encode the original intermediate representation to a lower space on mobile devices and reconstruct the representation at the edge. Also, knowledge distillation \cite{hinton2015distilling}, \gls{nas} \cite{ren2021comprehensive} and neural network pruning \cite{cheng2024survey} to manually create a bottleneck at the splitting point in order to reduce the dimensionality of latent representations. On the other hand, literature in \gls{sec} leverages different source coding approaches such as lossless coding (e.g. run-length coding \cite{carra2023dnn}, entropy coding \cite{matsubara2022supervised}) and lossy coding (e.g. JPEG \cite{lee2021splittable}, quantization \cite{matsubara2022bottlefit}) to further minimize the the average bit-length of latent representations. Recently inspired by the neural image compression \cite{balle2016end}, learning-driven rate-distortion optimization approaches are adopted to \gls{sec} \cite{matsubara2022supervised}. The key idea is to optimize \dnn based on the rate-distortion metric during training. After training, the learned latent representation can achieve minimum rate with entropy coding under certain distortion level. In addition, dynamic feature compression as an emerging technique has been studied recently to achieve wireless channel adaptive compression of latent representations \cite{abdi2023channel}.

\begin{wrapfigure}{l}{0.4\textwidth}
    \centering
    \includegraphics[width=\linewidth]{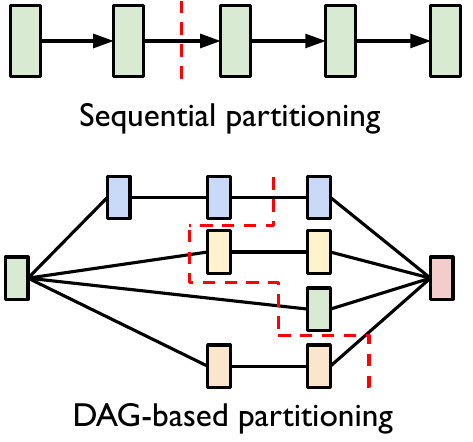}
    \caption{Difference of Sequential and DAG-based partition}
    \label{fig:partition}
\end{wrapfigure}

\subsection{Advancements in System Optimization} \label{sec:sys_opt}

Based on the type of partitioning graph, literature can be categorized as sequential-based partitioning and \gls{dag}-based partitioning. Figure~\ref{fig:partition} depicts the difference between them. In sequential-based partitioning, literature considers the \dnn as a Markov chain where the later layer only depends on the output of previous layer. However, modern \dnns such as Inception~\cite{szegedy2016rethinking} and Densenet~\cite{huang2017densely} are more complex, containing multiple path for the finer feature extraction. Simply selecting splitting point based on the sequential chain may not give optimal latency and energy consumption. Therefore, \gls{dag}-based partitioning is proposed to achieve a more granular \dnn splitting strategy by splitting the computing graph of the \dnn. Next, we review recent literature of sequential- and \gls{dag}-based approaches in resource-aware and dynamic model partitioning. Table~\ref{tab:so} gives a summary of the discussed work.

\subsubsection{Resource-Aware Partitioning}

\textbf{Sequential Partitioning.}~\citet{eshratifar2019jointdnn} investigated layer-wise splitting to minimize the energy consumption of mobile devices and end-to-end latency. It first profiled the energy consumption and execution latency of each \dnn layer during run-time. The optimal \dnn partitioning is then converted into integer linear programming problem and lagrange relaxation based method \cite{juttner2001lagrange} is used to constitute a lookup table for a different set of energy and networking constraints. Similarly, \citet{zawish2022towards} optimized the partitioning under various computing resource constraints (number of FLOPs), energy constraints and bandwidth constraints. \citet{gao2019deep} benchmarked layer-wise processing time on mobile and edge devices of different hyper parameters such as convolutional kernel size and data size. After benchmarking, a processing time estimation approach based on polynomial regression is developed. A mixed integer linear programming problem is formulated to minimize the end-to-end latency and solved with an opensource solver. \citet{gao2021task} further considered the \dnn partitioning in a multi-user scenario where multiple mobile devices will connect to the same edge device. The study jointly optimize the partitioning point and task scheduling to minimize inference latency, energy consumption and service fee paid to the server based on the aggregative game theory. \citet{ding2023resource} considered a privacy-preserving scenario for resource-limited \dnn partitioning. The proposed approach first collects per-layer computation and memory cost to build a set of valid splitting points. Then a distance correlation metric \citep{szekely2007measuring} is leveraged during training to reduce the information leakage of latent representations. \citet{li2024optimal} extended the \dnn partitioning to the multi-user cooperative inference system. The splitting and resource allocation are jointly optimized by transforming the problem into a mixed integer non-linear programming. A \gls{drl} algorithm is used to select splitting point to minimize energy consumption while a convex optimization algorithm is used for resource allocation. \rev{\citet{jiang2025energy} jointly optimized the partitioning and early exit of multi-branch \dnns to minimize the energy consumption. \citet{singhal2025distributing} focused on deploying multi-branch \dnns across mobile-edge-cloud systems. The optimization problem is modeled using a graph model to minimize the layer-to-layer energy cost. \citet{jung2025split} used a heuristic search to optimize the energy consumption and over-all latency in a multi-user scenario where multiple mobile devices and a single server performs continuous inference.} 

\textbf{DAG Partitioning.}~While above-mentioned literature considered sequential splitting, \citet{li2021throughput} investigated the \gls{dag}-based splitting and inference parallelism to accelerate the multiple inference requests. It involves partitioning the \dnn model and offloading parts to cloudlets while using multi-threading to meet inference delay requirements. \citet{duan2021computation} categorized \dnn graphs as tree-structure and path-independent structure and proposed different algorithms regarding the type of \gls{dag} to optimize end-to-end latency under resource constraints. Similarly, \citet{wang2021accelerate} partitioned \dnn based on its \gls{dag} structure to minimize latency. The problem is modeled with mixed integer programming which is proven np-hard and can be solved with heuristic search. \citet{duan2021joint} extended the partitioning problem for multiple \dnn tasks and proposed to minimize overall latency by jointly optimizing splitting point and task scheduling. \rev{\citet{liu2024adaknife} leveraged a greedy algorithm that can handle both chain and \gls{dag} structure to minimize the end-to-end latency.} \smallskip

\subsubsection{Dynamic Model Partitioning}

\textbf{Sequential Partitioning.}~Neurosurgeon \citep{kang2017neurosurgeon} is one of the earliest work considering dynamic model partitioning in different bandwidth and energy constraints. It first leverages a regression model to estimate the computing load (number of FLOPs) of different \dnn layers. Then, a greedy search approach is leveraged to select the splitting point in run-time to minimize the end-to-end latency or energy consumption of mobile devices. \citet{li2019edge} combined \dnn splitting with early exits and optimized the splitting point and exit point jointly in the changing networking environment where the bandwidth varies drastically based on the bayesian online change point detection \cite{adams2007bayesian}. \citet{mehta2020deepsplit} proposed to split \dnn dynamically based on the changing computation resources to minimize the overall transmission consumption. However, \cite{mehta2020deepsplit} does not consider the dynamic networking condition. \citet{karjee2021split} optimize the splitting point to minimize the computation latency in the dynamic computing resource constraints. In order to further optimize the communication latency, a dynamic network switching approach by leveraging both \gls{drl} and rule-based selection are proposed to change the wireless techniques such as bluetooth, cellular and Wi-Fi based on the network traffic. \citet{lim2022real} proposed to minimize the end-to-end latency using Lyapunov optimization. By modeling the dynamics in different time steps as a Lyapunov drift, \cite{lim2022real} utilized an real-time splitting approach to jointly determine the \dnn splitting point and the dynamic voltage and frequency scaling of the GPU on devices. \citet{bakhtiarnia2023dynamic} first profiled natural bottlenecks in off-the-shelf \dnns where the output is smaller than the original input data. Then, the \dnn is dynamically allocated on mobile and edge device aiming at minimizing the end-to-end latency. \citet{ko2023dynamic} investigated the dynamic model partitioning in the distributed severless edge scenario where different head and tail models are running in containers with different computing status. The optimization problem is modeled with a Markov decision process and linear programming is used to choose the optimal container to minimize the inference latency. \citet{nagamatsu2023dynamic} utilized mixed-precision \gls{nas} \citep{cai2020rethinking} to optimize the bit rate of latent representations for different workloads. As the bit rate can be dynamically changed, the \dnn is dynamically partitioned based on greedy search in order to minimize the inference time. \citet{lee2023wireless} proposed a two-stage wireless channel adaptive model partitioning addressing the energy and memory constraint of devices. The splitting point is first determined using an off-line exhaustive search without considering the wireless channel. After deployment, the model can be adjusted online to minimize inference latency according to wireless channel conditions. This approach reduced the search space in the online adjustment phase making it easy to deploy in real time. However, the limited adjustment points can result in sub-optimal latency minimization. \rev{\citet{liu2024dnn} leveraged multi-agent \gls{drl} to optimize the partitioning, task offloading and resource allocation in dynamic vehicular networks. Vehicles in the network act as independent agents to achieve distributed inference in a decentralized manner. \citet{qi2024multi} leveraged \gls{drl} to jointly optimize the model compression and model partitioning in varying network conditions. }

\textbf{DAG Partitioning.}~Along with chain-based splitting, literature also investigate \dnn partitioning based on their \gls{dag} structure. For example, \citet{hu2019dynamic} investigated the dynamic \gls{dag} partitioning in different \dnn workload scenarios to minimize the latency in low network traffic scenario and maximize the overall throughput in heavy network traffic scenario. \citet{zhang2021dynamic} introduced a dynamic \gls{dag} partitioning algorithm that can adjust the splitting point to its nearly vertex in the run time. \citet{liang2023dnn} further extended the work in \cite{hu2019dynamic} to a computing resource constrained scenario. \citet{li2024real} proposed an adaptive \gls{dag}-based splitting to optimize the latency and resource utilization in the multi-user scenario. \rev{\citet{nagamatsu2024mixed} applied \gls{nas} to optimize potential splitting points and dynamically selected splitting points based on the network condition.}

\begin{landscape}
\begin{table}
    \caption{Summary of Literature on System Optimization}
    \centering
    \resizebox{\linewidth}{!}{
    \begin{tabular}{cccccc}
    \toprule
        \textbf{Work} & \textbf{Application} & \textbf{Partition Type} & \textbf{Objective} & \textbf{Dynamic} \\
    \midrule
        \cite{eshratifar2019jointdnn} & classification, generative model & Sequential & minimize energy, latency & N/A \\
        \cite{zawish2022towards} & Image classification & Sequential & partitioning under various constraints & N/A \\
        \cite{gao2019deep} & Image classification & Sequential & minimize latency & N/A \\
        \cite{gao2021task} & Image classification & Sequential & minimize latency, energy and computing price & N/A \\
        \cite{ding2023resource} & Image classification & Sequential & preserve privacy under resource constraints & N/A \\
        \cite{li2024optimal} & \acrshort{csi} sensing & Sequential & minimize latency, energy consumption & N/A \\
        
        \rev{\cite{jiang2025energy}} & \rev{Image classification} & \rev{Sequential} & \rev{minimize energy consumption} & \rev{N/A} \\
        \rev{\cite{singhal2025distributing}} & \rev{Image classification} & \rev{Sequential} & \rev{minimize energy consumption, select early exit} & \rev{N/A} \\
        \rev{\cite{jung2025split}} & \rev{Image classification} & \rev{Sequential} & \rev{optimize latency and energy for multiple devices} & \rev{N/A} \\
        \rev{\cite{liu2024adaknife}} & \rev{Image classification} & \rev{Sequential and \gls{dag}} & \rev{minimize latency} & \rev{N/A} \\
        
        \cite{li2021throughput} & Image classification & \gls{dag} & maximize throughput for parallel tasks & N/A \\
        \cite{duan2021computation} & Image classification & \gls{dag} & minimize latency & N/A \\
        \cite{duan2021joint} & Image classification & \gls{dag} & minimize overall latency for multiple \dnn inference & N/A \\
        \cite{wang2021accelerate} & Image classification & \gls{dag} & minimize latency & N/A \\
        \cite{kang2017neurosurgeon} & Image classification & Sequential & minimize latency, energy consumption & Yes \\
        \cite{li2019edge} & Image classification & Sequential & minimize latency; preserve inference accuracy & Yes \\
        \cite{mehta2020deepsplit} & Object detection & Sequential & minimize bandwidth occupation & Yes \\
        \cite{karjee2021split} & Image classification & Sequential & minimize latency; optimal network switching & Yes \\
        \cite{lim2022real} & Object detection & Sequential & minimize latency, energy consumption & Yes \\
        \cite{bakhtiarnia2023dynamic} & Image classification & Sequential & minimize latency & Yes \\
        \cite{ko2023dynamic} & Image classification & Sequential & minimize latency; maintain computing status & Yes \\
        \cite{nagamatsu2023dynamic} & Image classification & Sequential & minimize latency; minimize datasize & Yes \\
        \cite{lee2023wireless} & Image classification & Sequential & minimize latency & Yes \\

        \rev{\cite{liu2024dnn}} & \rev{Traffic sign detection} & \rev{Sequential} & \rev{minimize latency} & \rev{Yes} \\
        \rev{\cite{qi2024multi}} & \rev{Image classification} & \rev{Sequential} & \rev{balance latency and model compression} & \rev{Yes} \\
        
        \cite{zhang2021dynamic} & Image classification & \gls{dag} & minimize latency & Yes \\
        \cite{hu2019dynamic} & Video analysis & \gls{dag} & minimize latency; maximize throughput & Yes \\
        \cite{liang2023dnn} & Video analysis & \gls{dag} & minimize latency; maximize throughput & Yes \\
        \cite{li2024real} & Image classification & \gls{dag} & minimize latency; resource allocation & Yes \\
        
        \rev{\cite{nagamatsu2024mixed}} & \rev{Image classification} & \rev{\gls{dag}} & \rev{minimize latency} & \rev{Yes} \\
    \bottomrule
    \end{tabular}
    }
    \label{tab:so}
\end{table}
\end{landscape}

\subsubsection{\rev{Research Challenges}} \label{sec:challenge1}

\rev{\textbf{Model Dependencies.}~One significant challenge in system optimization is finding the accurate computation characteristics--such as latency, energy consumption, and computing memory--of each possible partitioning point. Most existing literature relies on an exhaustive performance analysis for every layer in \dnns \cite{eshratifar2019jointdnn,gao2019deep,gao2021task,ding2023resource}. While this approach provides accurate profiling, it makes the optimization model-specific. For different \dnn architectures such as CNNs and transformers that have distinct computation and memory usage patterns, even when they have the same computation complexity (number of FLOPs), the latency can be significantly different. This makes optimization algorithms designed for a specific model potentially non-generalizable to others.} \smallskip

\rev{\textbf{Hardware Dependencies.}~Hardware with different computing capabilities can produce varying latency for identical computation patterns \citep{qin2024mobilenetv4}. Only profiling layer-wise execution of \dnns on a single device may fail to generalize to heterogeneous platforms. different platforms often use different DNN compilers (e.g., TensorFlow, PyTorch, ONNX) with varying compatibility requirements. Creating partitioning frameworks that can effectively accelerate DNNs across heterogeneous infrastructures presents significant research challenges \cite{liu2024adaknife}. } \smallskip

\rev{\textbf{Optimization Complexity.} The search space for optimal partitioning is growing exponentially with an increasing \dnn model complexity (e.g., \dnns with hundreds of layers), making it computationally intractable to evaluate all possible configurations. Approaches that rely on heuristics or approximation algorithms \cite{wang2021accelerate,jung2025split,nagamatsu2023dynamic} can reduce the complexity but remain challenging in finding globally optimal solutions. Complex optimization algorithms \cite{li2024optimal,hu2022distributed,gao2021task} provide better optimized performance but result in high latency, preventing real-time adaptation in dynamic environments.} \smallskip

\rev{\textbf{Reconfiguration Cost.} Network conditions, hardware availability, and workload characteristics change during distributed inference. Static partitioning decisions made at deployment time often become suboptimal, but dynamic repartitioning introduces challenges in maintaining hardware computing status \cite{ko2023dynamic}. In addition, most dynamic optimization methods work in a centralized manner \cite{kang2017neurosurgeon,nagamatsu2023dynamic,mehta2020deepsplit}. It requires additional orchestration to synchronize the deployment on both mobile and edge devices which result in extra communication and computation overhead. }

\subsection{Advancements in Semantic Compression} \label{sec:semcomp}

Based on techniques utilized in literature, semantic compression can be categorized as dimension reduction, bit-width reduction and learning-driven rate-distortion optimization approaches. Table~\ref{tab:sc} gives a brief summary of recent progress in semantic compression. Next, we dive into the technical details of these approaches.

\subsubsection{Dimension Reduction}

\begin{figure}[!t]
    \centering
    \includegraphics[width=0.95\linewidth]{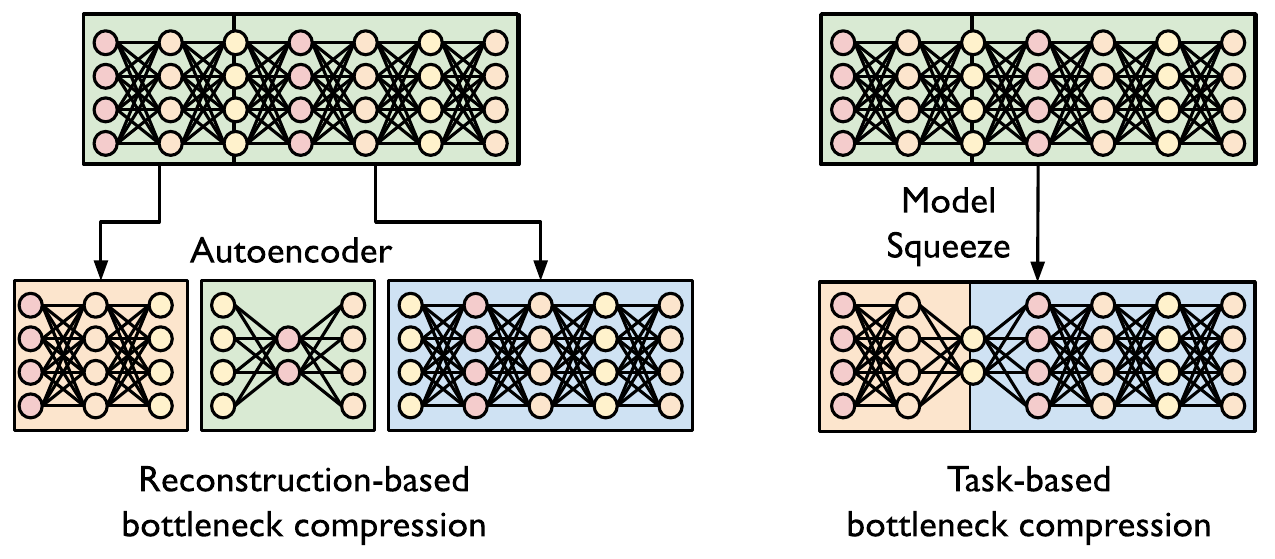}
    \caption{Two bottleneck compression methodologies.}
    \label{fig:dim_comp}
\end{figure}

Early work in \gls{sec} splits pre-trained \dnns without modifying model architectures \citep{kang2017neurosurgeon,eshratifar2019jointdnn,li2018learning,jeong2018computation}. However, as off-the-shelf \dnns are not designed for collaborative inference, they often use large latent dimensions in early layers to capture redundant features to enhance end-to-end performance. Therefore, naively splitting \dnn may incur excessive communication overhead due to the dimension of semantics \citep{shao2020communication}. To reduce the communication overhead in \gls{sec}, feature compression layers (i.e., ``bottleneck'') is introduced to squeeze the dimension of latent representations \citep{hu2020fast,yao2020deep,hao2022multi,matsubara2020head,matsubara2022bottlefit}. As depicted in Figure~\ref{fig:dim_comp}, there are two major bottleneck compression approaches categorized as reconstruction-based and task-oriented bottleneck. Similar to lossy data compression studies \cite{cheng2018deep,liu2021high}, reconstruction-based bottleneck introduces an auto-encoder at the splitting point, aiming at compressing and reconstructing the latent representations by minimizing the distance metric (e.g. MSE loss) between the original and reconstructed representations. For example, \citet{hu2020fast} injected a symmetric auto-encoder at the \dnn splitting point and train the whole model in an end-to-end manner. Similarly, \citet{hao2022multi} used an auto-encoder to reduce the dimension in multi-agent collaborative tasks. The auto-encoder is trained separately to reconstruct the latent representation. \citet{yao2020deep} introduced an imbalanced encoder-decoder structure by taking the mobile resource constraints into account. The encoder on the mobile device has a light-weight design while the decoder on the edge device has more layers for better feature reconstruction.

On the other hand, inspired by \dnn compression, the task-oriented bottleneck proposes to directly squeeze the network architecture and train the modified model with task-oriented loss (e.g. cross-entropy loss). Model compression approaches such as knowledge distillation \cite{hinton2015distilling}, \gls{nas} \cite{ren2021comprehensive} and \dnn pruning \cite{cheng2024survey} are often leveraged to achieve superior compression performance while preserving accuracy. For example, \citet{eshratifar2019bottlenet} modified the size of the early layers in the \dnn to decrease computational load on mobile devices. The bottleneck is introduced in the middle of the modified architecture and the model is trained with cross entropy loss. However, it suffers performance loss as introducing bottleneck in early layers can result in loss of task-oriented information. As such, \citet{matsubara2021neural} proposed to train the bottleneck using knowledge distillation. Similarly, \citet{dong2022splitnets} optimized the feature compression design using \gls{nas} while \citet{guo2023digital} utilized pruning to reduce the bottleneck size. \rev{\citet{mendula2025novel} use knowledge distillation and FP16 quantization to reduce the communication overhead for object detection tasks.}

\begin{table}[!t]
    \centering
    \caption{Summary of Literature on Semantic Compression}
    \resizebox{\linewidth}{!}{
    \begin{tabular}{cccc}
    \toprule
    \textbf{Work} & \textbf{Dimension Reduction} & \textbf{Bit-Rate Reduction} & \textbf{R-D Optimization} \\
    \midrule
    \cite{hu2020fast} & auto-encoder & N/A & No\\
    \cite{dong2022splitnets} & \gls{nas} & N/A & No\\
    \cite{yao2020deep} & auto-encoder & Entropy coding & No\\
    \cite{hao2022multi} & auto-encoder & 8-bit quantization & No\\
    \cite{eshratifar2019bottlenet} & auto-encoder & 8-bit quantization and JPEG & No\\
    \cite{lee2021splittable} & auto-encoder & JPEG & No\\
    \cite{li2023attention} & auto-encoder & Entropy coding & No\\
    \cite{matsubara2020head} & Distillation & 8-bit quantization & No\\
    \cite{matsubara2022bottlefit} & Distillation & 8-bit quantization & No\\
    \cite{matsubara2021neural} & Distillation & 8-bit quantization & No\\
    \rev{\cite{mendula2025novel}} & \rev{Distillation} & \rev{FP16 quantization} & \rev{No} \\
    \cite{guo2023digital} & Pruning & Non-linear quantization  & No\\
    \cite{carra2023dnn} & N/A & Run-length coding & No\\
    \cite{ko2018edge} & N/A & JPEG; Run-length and Huffman coding & No\\
    \cite{cohen2021lightweight} & N/A & Entropy coding & No\\
    \cite{chen2019toward} & N/A & Various lossless and lossy coding & No\\
    \cite{wang2022channel} & N/A & minimize bitwidth with greedy search & No\\
    \cite{alvar2021pareto} & N/A & minimize bitwidth with convex opt. & No\\
    \rev{\cite{xiao2024adaptive}} & \rev{N/A} & \rev{adaptive bitwidth based on attn map} & \rev{No} \\
    \rev{\cite{li2024dynamic}} & \rev{auto-encoder} & \rev{adaptive bitwidth and entropy coding} & \rev{No} \\
    \cite{matsubara2022sc2} & distillation & entropy coding & Yes\\
    \cite{matsubara2022supervised} & distillation & entropy coding & Yes\\
    \cite{duan2022efficient} & distillation & entropy coding & Yes\\
    \rev{\cite{harell2025rate}} & \rev{distillation} & \rev{entropy coding} & \rev{Yes} \\
    \cite{yamazaki2022deep} & reconstruction & entropy coding & Yes\\
    \cite{datta2022low} & reconstruction & entropy coding & Yes\\
    \cite{ahuja2023neural} & reconstruction & entropy coding & Yes\\
    \cite{hossain2023flexible} & reconstruction & entropy coding & Yes\\
    \rev{\cite{yuan2024split}} & \rev{reconstruction} & \rev{entropy coding} & \rev{Yes} \\
    \bottomrule
    \end{tabular} }
    \label{tab:sc}
\end{table}

\subsubsection{Bit-Rate Reduction} 

Source coding is an effective way to compress the information in original data with fewer bits. For example, \citet{carra2023dnn} argued that a carefully designed linear quantization and run-length coding approach can achieve similar compression performance in comparison to bottleneck approaches. Therefore, various bit-rate reduction technique are leveraged in literature for semantic compression. Based on the source coding technique, literature can be categorized with lossless coding (e.g. run-length coding, entropy coding) and lossy coding (e.g., JPEG, quantization). For example, \citet{chen2019toward} benchmarked various existing lossless (e.g. GZIP \cite{deutsch1996gzip}; ZLIB \cite{gailly2004zlib}; BZIP2 \cite{seward1996bzip2}; and LZMA \cite{louchard1997average}) and lossy compression (HEVC \cite{sze2014high}) techniques on intermediate representations for different \dnn architectures. \citet{ko2018edge} applied run-length and Huffman coding to achieve 3-10x compression rate and JPEG to achieve up to 50x compression rate without bottleneck. However, simply applying lossless coding can achieve limited compression performance. To achieve a significant compression rate, \citet{cohen2021lightweight} introduced a light-weight compression method for pre-trained object detectors based on clipping, binarization and entropy coding. \citet{wang2022channel} proposed a multi-stage quantization algorithm based on feature statistics and greedy search to minimize the information loss in bit-rate reduction. \citet{alvar2021pareto} studied the optimal bit allocation of intermediate feature for multi-stream tasks by modeling it as a multi-objective optimization problem. \rev{\citet{xiao2024adaptive} proposed a block-wise variable bit-rate quantization algorithm based on the attention map weight in transformers.}

Note that bit-rate reduction can be applied on top of dimension reduction approaches such as bottleneck compression \citep{yao2020deep,hao2022multi,matsubara2022bottlefit}. For example, \citet{eshratifar2019bottlenet} used JPEG to further compress the latent features after dimension reduction for image classification tasks. As the JPEG compressor is non-differentiable, an identical mapping is used to replace the compressor during backpropagation in order to train the model. \citet{lee2021splittable} leveraged binarization and PNG encoding to reduce the size of bottleneck output with only 1\% performance loss for object detection tasks compared to YOLOv5 \citep{Jocher_YOLOv5_by_Ultralytics_2020}. \citet{li2023attention} first leveraged channel-wise attention \citep{woo2018cbam} to prune the auto-encoder and then applied entropy coding to reduce the bit-rate of the compressed semantic. \citet{guo2023digital} proposed to use pruning to reduce the dimension and a learnable non-linear quantization block \citep{yang2019quantization} to compress the semantic with an adaptive bit width. \rev{\citet{li2024dynamic} used auto-encoder to compress the dimension and applied a variable depth quantization based on the attention map of CNN channels. Entropy coding is further leveraged to compress the datasize.}

\subsubsection{Rate-Distortion Optimization} 
While bit-rate reduction has been proven to be effective, the latent representation may not be optimal for source coding. Thus, recent work proposed to improve the compression performance by regularizing the learned semantic with the rate-distortion metric. Based on Shannon's rate-distortion theorem \citep{cover1999elements}, the rate-distortion function is defined as the minimum achievable rate $R$ under a distortion constraint $D$. Therefore, the compression rate can be optimized by the following objective:
\begin{equation}
    \min \quad R + \beta D \label{r-d_tradeoff}
\end{equation}
where $\beta$ is a Lagrangian multiplier to control the rate-distortion trade-off.

\begin{wrapfigure}{l}{0.5\linewidth}
    \centering
    \includegraphics[width=\linewidth]{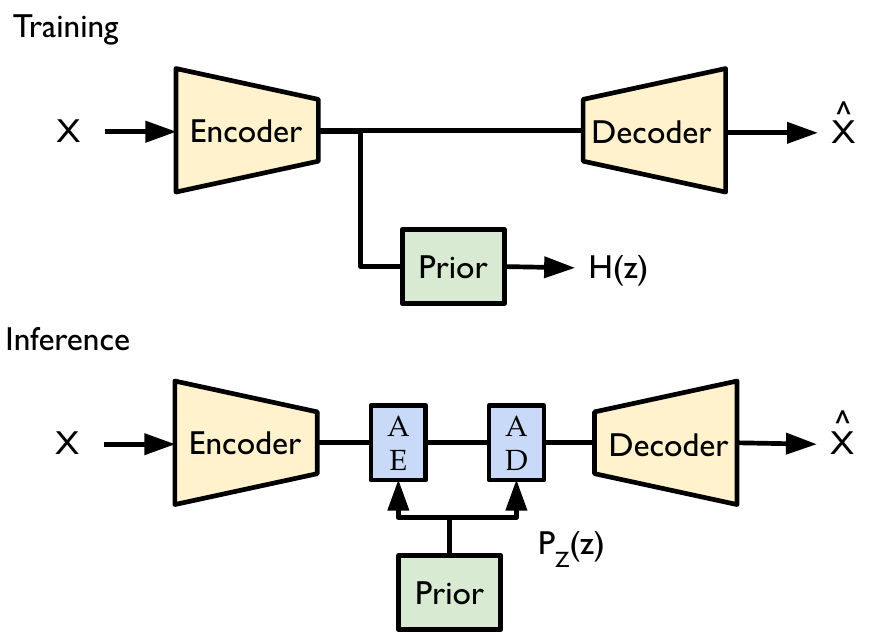}
    \caption{Overview of learning-driven entropy coding.}
    \label{fig:entropy}
\end{wrapfigure}

Inspired by the rate-distortion optimization, \citet{balle2016end} first proposed a neural image compression method based on a \gls{vae} \citep{kingma2013auto} and a factorized prior. A general framework of learning-driven entropy coding is depicted as Figure~\ref{fig:entropy}. A \gls{vae} is used to compress and quantize the image to latent representation while a prior model is used to estimate the entropy of the compressed information. According to Shannon's Coding Theorem \citep{cover1999elements}, the lossless coding rate of a message is lower bounded by its entropy. Therefore, the objective of the image compression is to optimize the following rate-distortion trade-off:
\begin{equation}
    \min \quad \underset{rate}{\underbrace{H(z)}} + \underset{distortion}{\underbrace{\beta D(x,\hat{x})}} \label{rd_image}
\end{equation}
where $H(z)$ denotes the entropy of latent representation $z$ and $D(x,\hat{x})$ is the distortion between original image $x$ and reconstructed image $\hat{x}$. After training, estimated entropy is used to encode latent representation with lossless coding algorithm such as arithmetic coding \citep{rissanen1981universal}. 

While entropy-based optimization approach is originally proposed for image compression, it is adopted to semantic compression rapidly. For example, \citep{duan2022efficient} applied the entropy-based compression before the pre-trained \dnn to directly compress the raw input. As the pre-trained \dnn is not optimized for reconstructed input, it requires a multi-stage training. The first stage is to train the compressor with Eqn.~\ref{rd_image} and the second stage is to fine-tune the \dnn with task-oriented metrics (e.g. cross-entropy loss). In addition to add compression before pre-trained \dnns, other work injects entropy-based compression in the middle of \dnns. For example, in \citep{yamazaki2022deep}, an auto-encoder with factorized prior \citep{balle2016end} is trained to reconstruct the latent representation of pre-trained \dnns. \citet{datta2022low} proposed a similar approach but with a different entropy estimation module \citep{theis2017lossy}. \citet{ahuja2023neural} injected the entropy compressor to compress the semantic and applied knowledge distillation to fine-tune the performance of edge \dnn. \citet{hossain2023flexible} does not optimize to reconstruct the latent representation but replace the distortion $D$ in Eqn.~\ref{rd_image} with cross-entropy loss. To reduce the computation overhead introduced by entropy-based compression, \citet{matsubara2022supervised} directly applied the rate-distortion optimization to exisiting light-weight bottleneck \citep{matsubara2022bottlefit} and the multi-stage training approach in \citep{matsubara2022bottlefit} is also applied to preserve the task-oriented performance. \citet{matsubara2022sc2} systematically benchmarked the compression performance of \citep{matsubara2022supervised} and other dimension or bit-rate reduction approaches. Experiments demonstrated that the learnable entropy coding can achieve the best compression-computation-performance trade-off in comparison to dimension reduction and naive bit-rate reduction. \rev{\citet{harell2025rate} systematically investigated the theory and designing strategies including the splitting point choice, training strategy and architecture design of learning-driven rate-distortion optimization in split computing. \citet{yuan2024split} created layers with different dimensions to achieve an adaptive compression performance and proposed a multi-round refinement method to optimize the rate-distortion. }

\subsubsection{\rev{Research Challenges}} \label{sec:challenge2}

\rev{\textbf{Bottleneck Optimization.} Injecting bottleneck in \dnns to reduce data dimension can jeopardize the task-oriented performance. To preserve the accuracy, research usually involves advanced training methods to optimize the bottleneck. For example, \citet{yao2020deep} applied multi-stage training to auto-encoder. In the first stage, the auto-encoder is trained to reconstruct the semantic features and in the second stage, knowledge distillation is applied to fine-tune the auto-encoder. \citet{hao2022multi} add a performance regularization term to reconstruction loss when training the auto-encoder. After training the auto-encoder, the partitioned \dnn on the edge device is fine-tuned to further recover the performance. \citet{matsubara2022bottlefit} first applied knowledge distillation to train the encoder and then fine-tune the decoder to further mitigate the performance drop. } \smallskip

\rev{\textbf{Source Coding Complexity.} To achieve better data compression performance, literature usually leverage complex optimization approaches to find the optimal bit-rate allocation for different semantic features \citep{alvar2021pareto,wang2022channel,carra2023dnn}. However, optimization-based source coding can introduce additional latency during inference. On the other hand, simple coding scheme such as JPEG and uniform bit quantization may result in information loss and less optimal compression rate compared to others \citep{matsubara2022sc2}. } \smallskip

\rev{\textbf{Resilience of Entropy Models.}~Learning-driven rate-distortion methods can jointly optimize the bottleneck and source coding \cite{ahuja2023neural,harell2025rate}, resulting in a better compression performance compared to other methods separately optimize bottleneck and source coding \cite{matsubara2022bottlefit,eshratifar2019bottlenet}. However, as the entropy estimation module is learned over a specific dataset, a distribution shift in data source may drastically affect the source coding rate. \citet{zhang2024resilience} demonstrated that the compression performance of learned entropy model is vulnerable to interference such as short noise and adversarial attack. Therefore, the major challenge of entropy-based rate-distortion optimization is to train a compression module that is resilient to intentional and unintentional noise.}

\section{Recent Progress in Semantic Communication} \label{sec:semcomm}

\subsection{Taxonomy of Semantic Communication}

\begin{wrapfigure}{l}{0.5\linewidth}
    \centering
    \includegraphics[width=\linewidth]{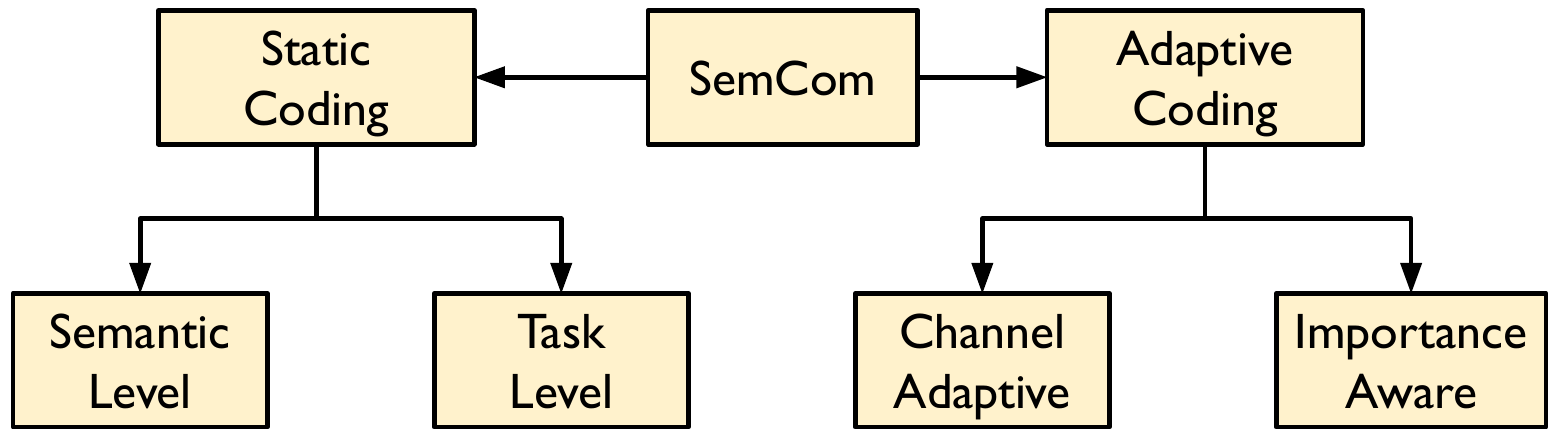}
    \caption{Taxonomy of Recent Progress in Semantic Communication}
    \label{fig:t-semcom}
\end{wrapfigure}

The taxonomy of \gls{semcom} is shown in Figure~\ref{fig:t-semcom}. We classify recent advancements in \gls{semcom} into two categories: train-time static coding and test-time adaptive coding. Static \gls{semcom} systems focus on generating semantic representations that are resilient to stochastic wireless channels, and can be further divided into semantic-level coding and task-level coding. On the other hand, adaptive \gls{semcom} systems aim to generate semantics dynamically in response to varying channel conditions, which can be further categorized into channel-adaptive coding and importance-aware coding. Table~\ref{tab:semcom} summarizes the recent literature reviewed in this paper.

\subsubsection{Train-Time Static Systems}

The existing static methods to develop \gls{semcom} systems take the wireless channel effects into account when end-to-end training the \dnn by adopting a stochastic model for the channel. The channel model should be differentiable to enable updating the encoder \dnn parameters using back-propagation. Benefiting from their data-driven nature, static \gls{semcom} systems show promising performance in complicated communication scenarios where it is difficult to describe the system with mathematical models or an analytically-optimal solution cannot be derived. However, the transmitter and receiver \dnns are fixed after training in such systems, thus sending a constant amount of data regardless of the input, communication environment conditions, and user preferences. This necessitates an accurate modeling of the channel-related distortions through a differentiable stochastic model.

Early work \cite{o2017introduction, bourtsoulatze2019deep} in this domain assumed a simple \gls{awgn} model for the wireless channel which is implemented using a non-trainable noise-injection layer and added between the encoder and decoder \dnns during training. Such a model fails to express the real-world channel characteristics which in turn limits the performance of such \gls{semcom} systems after deployment. \rev{Other work \cite{xie2021deep, kurka2020deepjscc, xie2020lite, zhu2024evaluate} proposed different semantic similarity metrics or more expressive architectures for the \dnn to help improve the trained system generalization ability to real-world channels.} To support multi-path fading effects, \citet{felix2018ofdm, o2017physical} employ Raleigh and Rician models for channel impairments. However, due to the mismatch between the stochastic channel model used for training and the actual channel transfer function during test time, real-world implementations suffer from a noticeable performance drop compared to the results obtained during simulation.

\subsubsection{Test-Time Adaptive Systems}

A major challenge in developing static \gls{semcom} systems is that a constant stochastic model which can represent the channel accurately enough is often hard to obtain in training phase. Also, the stochastic model chosen to represent the channel may not be differentiable. Therefore, existing studies resort to either approximating the wireless effects by a parameterized differentiable model or estimating the instantaneous channel gradients to enable the dynamic optimization of the encoder. For example, \citet{ye2020deep, o2019approximating} proposed to train a \gls{gan} as a surrogate \dnn model for the channel. Also, \citet{raj2018backpropagating} proposed perturbing the symbols to obtain an estimate of the instantaneous channel derivatives instead of the full channel estimation. This allows updating the encoder \dnn parameters through back-propagation.

On the other hand, as the channel changes over time due to the stochastic variations in the surrounding environment, other adaptive systems are proposed which leverage test-time adaptation techniques to update the encoder, the decoder, or both. For example, \citet{aoudia2019model} trained a \gls{drl} agent that can update the encoder \dnn weights to optimize the final metrics. Also, \citet{park2020end} utilized meta-learning to train a module which can quickly adapt the decoder to the ongoing channel conditions.

While in these methods the architecture of the encoder and decoder \dnns are fixed and only their parameters are updated, recent work \cite{abdi2023channel} use dynamic pruning to change the number of transmitted symbols based on the ongoing wireless link quality. Moreover, other work \cite{huang2020clio} focuses on making the \gls{semcom} systems adaptive to the data rather than the channel. For example, the authors in \cite{huang2020clio} extract the importance score of different semantics and prioritize the transmission of the most important ones in severe channel conditions.

\subsection{Advancements in Train-Time Static Systems}

\subsubsection{Semantic-Level Coding}

\citet{o2017introduction} first proposed using an auto-encoder with a noise-injection layer to implement an end-to-end communication system assuming an \gls{awgn} channel model. More specifically, the encoder and decoder part of the auto-encoder are deployed at the transmitter and receiver, respectively, while the non-trainable noise injection layer is employed only during training for the \dnn to learn the channel distortion patterns. \citet{o2017deep} extended this idea to account for more realistic channel models such as Rayleigh multi-path fading. Also, \citet{erpek2018learning} improved the study by developing a \gls{phy} scheme for \gls{mimo} systems in the presence of interference. All of these methods obtained superior performance compared to classical modular communication system designs.

\begin{wrapfigure}{l}{0.5\linewidth}
    \centering
    \includegraphics[width=\linewidth]{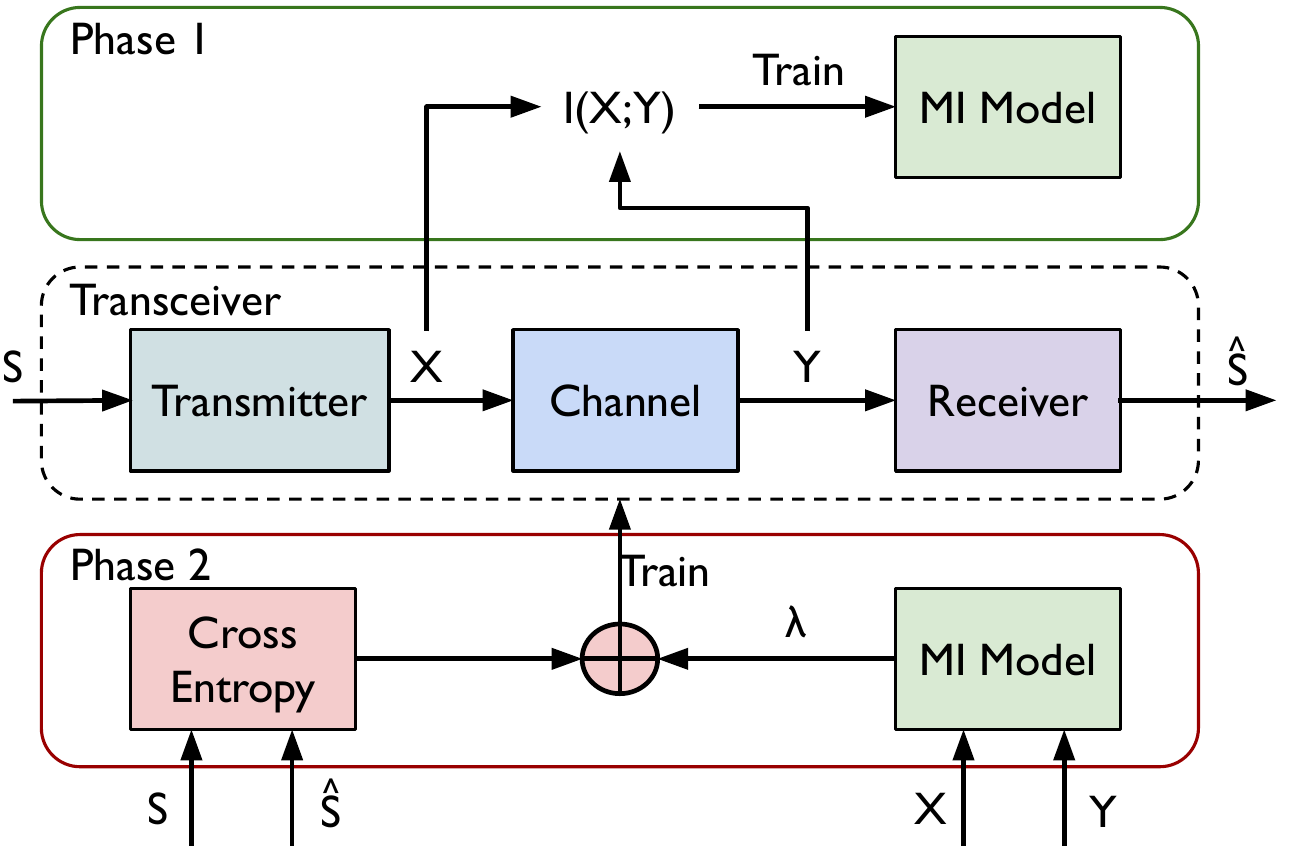}
    \caption{The training framework of the DeepSC \cite{xie2021deep}.}
    \label{fig:deepsc}
\end{wrapfigure}

\textbf{Text transmission.}~The aforementioned \gls{dl}-based communication systems laid the foundation for the development of \gls{semcom}. As such, \glspl{lstm} are used in \cite{farsad2018deep} to perform \gls{jscc} for text transmission. The proposed method is capable of preserving semantic information by mapping the semantically similar sentences onto closer points in the embedding space and using the \gls{wer} instead of \gls{ber} to train the \dnn. Having a similar design to the second level of Figure~\ref{fig:semcom}, \citet{xie2021deep} (DeepSC) used transformers \cite{subakan2021attention} both as the semantic encoder and decoder and added dense layers on top of them for channel coding which facilitates subsequent transmission. They developed a loss function $L = CE(S, \hat{S}) - \lambda I(X; Y)$ to train the entire system, where the first term minimizes the cross-entropy (semantic difference) between $S$ and $\hat{S}$ and the second term maximizes robustness to channel distortions. As illustrated in Figure~\ref{fig:deepsc}, the training framework of DeepSC has two stages: In phase 1, the mutual information estimation model is trained, and in phase 2, the whole system is trained considering both the cross entropy and the mutual information loss.

A lightweight version of DeepSC is developed in \cite{xie2020lite}, where the authors prune the \dnn and reduce the resolution of its parameters to obtain an \gls{iot}-friendly model. \citet{sana2022learning} adopted a novel loss function $L = I(X; S) - (1 + \alpha) I(X; Y) + \beta KL(Y, \hat{S})$ by taking into account the compactness of the extracted semantics. In this loss function, the first two terms are used to ensure compactness and robustness respectively, while the last term minimizes the semantic distortion incurred during transmission. Moreover, the authors proposed a semantic-adaptive mechanism which dynamically adapts the number of symbols per word to achieve a better trade-off between performance and communication efficiency.

\textbf{Image transmission.}~\Gls{jscc} schemes have been employed for image transmission which directly map raw pixel values to complex-valued channel symbols. In these schemes, the \gls{dl}-based semantic encoder first extracts low-dimensional semantic representations which are then distorted according to a channel model. \rev{The semantic decoder recovers the images based on the received symbols, where the end-to-end system is trained to maximize the semantic similarity metrics between the original and recovered images such as \gls{psnr}, \gls{ssim}, and \gls{lpips}~\cite{zhang2018unreasonable}}. \citet{bourtsoulatze2019deep} first proposed a technique to transmit images called DeepJSCC, where a \gls{cnn}-based autoencoder is used as the \gls{jscc} model. \citet{kurka2020deepjscc} introduced the variable-length coding feature by utilizing channel feedback (i.e., DeepJSCC-f), where the transmission of each image is done sequentially in $L$ stages. Specifically, each stage improves the quality of the recovered image by sending additional semantics corresponding to the residual error from the previous stage. \citet{kurka2021bandwidth} redesigned a DeepJSCC framework for bandwidth-adaptive transmission, where images are transmitted progressively in layers over time. As such, they proposed a successive refinement scheme for \gls{jscc} where the receiver can aggregate the received layers sequentially to increase the quality of the reconstructed semantics. \citet{zhang2022wireless} proposed a multi-level semantic feature extractor for image transmission, called MLSC-image, which encodes both the high-level information such as text and segmentation semantics and the low-level information such as local spatial details. These extracted features are later combined and encoded into channel symbols. \rev{To make \gls{jscc} systems compatible with digital modulation schemes, \citet{bo2024joint} developed a joint coding-modulation framework, called JCM, based on \glspl{vae} that outputs discrete constellation symbols as the encoded representation. The presented framework is trained in an end-to-end fashion to match the obtained modulation strategy with the operating channel condition.} \rev{The authors in \cite{park2024joint} improved the robustness of digital \gls{semcom} systems by developing a demodulation method that can assess the uncertainty of the demodulation output. Moreover, they considered a binary symmetric erasure model for the channel and proposed adaptively determining the modulation order based on channel conditions.}

\rev{Generative \gls{ai} has been leveraged in \cite{zhang2025semantic} to increase the perceptual quality of the reconstructed images in \gls{semcom} systems. More specifically, the source and channel coding mechanism at the transmitter end is implemented by a Swin Transformer, while a diffusion model is used at the receiver to decode high-quality images.} \rev{Also, \citet{wu2024cddm} proposed using a diffusion model to gradually remove the channel noise in \gls{semcom} systems. To achieve this, the diffusion model is applied after the channel equalization module and learns the distribution of the channel input signal.} \rev{\citet{tang2024contrastive} utilize contrastive learning to develop a semantic contrastive coding scheme. In the proposed scheme, input images and their reconstructed outputs are projected onto a latent space, where the projections of similar samples are pulled together, and the projections of irrelevant inputs are pushed apart following contrastive learning ideas. Moreover, when a \dnn for the downstream task is not available, a duplicate of the semantic encoder is deployed at the receiver to facilitate the training process.}

While the aforementioned studies have differentiable objectives to optimize, other methods focus on utilizing \gls{drl} to address user-defined non-differentiable semantic metrics which commonly arise in real-world wireless systems. \citet{lu2021reinforcement} first proposed a \gls{drl}-based \gls{semcom} system, called SemanticRL-JSCC, where a self-critic policy is learned to maximize the semantic similarity measures such as \gls{bleu} \cite{papineni2002bleu} and \gls{cider} \cite{vedantam2015cider}. Furthermore, to address non-differentiable channels, SemanticRL-JSCC learns separate policies for the encoder and decoder which shows strong robustness under fluctuating \glspl{snr}. \citet{lu2022rethinking} also used \gls{drl} to develop a joint semantics-noise coding (JSNC) method capable of adapting to time-varying channels while ensuring the preservation of semantic information. As such, a confidence-based distillation mechanism is employed in \cite{lu2022rethinking} both at the encoder and decoder which ensures the confidence reaches a threshold before the encoder/decoder could further process the semantics. \citet{huang2022toward} proposed a RL-based adaptive semantic coding (RL-ASC) technique which allocates different quantization levels for different semantic concepts based on their importance. Specifically, the crucial semantics are encoded with a higher precision to minimize a triple-term (rate, semantic, and perceptual) loss, while the task-irrelevant concepts can be discarded. Apart from the RL-based bit allocation module, \cite{huang2022toward} used a convolutional semantic encoder, and a \gls{gan}-based decoder equipped with an attention module which can fuse both local and global features.

\rev{Some existing work utilizes multi-modal features to improve the transmission quality of \gls{semcom} systems. For example, \citet{zhao2024lamosc} propose a \gls{semcom} system (LaMoSC) which integrates both the visual and textual features extracted from input images using an incorporated pre-trained \gls{llm}.} \rev{The authors in \cite{jiang2024large} propose using a multi-modal language model to align the semantics in image, audio, video, and text data, and then utilize a personalized \gls{llm} as the knowledge base both at the transmitter to extract the semantics and at the receiver to recover the messages. Moreover, a \gls{cgan} is employed for channel estimation based on the transmitted and received pilot signals.}

\subsubsection{Task-Level Coding}

While \gls{sl}-\gls{semcom} systems aim to reconstruct the semantics of the source message at the receiver side, \gls{tl}-\gls{semcom} frameworks are developed to enable the source and destination to effectively and efficiently execute a task. As such, \gls{tl}-\gls{semcom} systems require to transmit only the task-specific semantics instead of all the semantics which allows them to achieve higher communication efficiency compared to their \gls{sl}-\gls{semcom} counterparts. The underlying concepts in this line of research are therefore similar to Distributed \dnns. The major difference between these two research fields is that while in Distributed \dnns mobile devices use full-stack network protocols such as Wi-Fi or cellular networks to transfer the compressed semantics to the edge server, \gls{tl}-\gls{semcom} systems also perform channel coding and transmit raw \gls{phy} symbols over the channel. In addition, \gls{tl}-\gls{semcom} work often incorporate the channel model into the end-to-end training of the \dnn which makes the encoded semantics robust to channel distortions. Overall, even though Distributed \dnn and \gls{tl}-\gls{semcom} are considered two sides of one coin, the literature in the former area is focused on the compression part, while the latter mostly focuses on the communication aspects.

While \gls{jscc} schemes in \gls{sl}-\gls{semcom} train the encoder and decoder \dnns using semantic similarity metrics, \citet{strinati20216g} proposed a task-oriented \gls{jscc} approach by replacing the decoder output with the task inference result and using a task-specific loss to train the \dnns instead. \citet{jankowski2020deep, jankowski2020wireless} studied an image retrieval task and used fully-connected layers for \gls{jscc} encoder and decoder. Also, \citet{shao2020bottlenet++} added convolutional layers for channel coding between the semantic encoder and decoder to perform task-oriented \gls{jscc} for an image classification task. \citet{kang2022task} considered a scene classification task and proposed a \gls{drl}-based algorithm to find the most task-relevant semantic blocks. The \gls{drl} policy can strike a balance between communication latency and task performance by removing the inessential blocks according to the observed channel conditions.

\citet{shao2021learning} utilized the Information Bottleneck (IB) theory to induce sparsity in the channel-encoded semantics. Such sparsity is further used to prune the redundant dimensions of such representation, thus reducing the transmitted data size. However, the transmitted representations in \cite{shao2021learning} are continuous-valued which require an infinite-resolution constellation design for the adopted modulation scheme \cite{wang2022constellation, dai2022nonlinear}. \citet{xie2023robust, abdi2025phydnns} resolved the compatibility issue with digital modulation-based communication systems by proposing a discrete task-oriented \gls{jscc} where the encoded representations are discretized using a Gumbel-based vector quantization block. \rev{As such, their method is the task-oriented equivalent of the semantic-level \gls{semcom} system presented in \cite{bo2024joint}.}

The multi-modal data transmission has also been investigated under various settings. \citet{xie2021task} proposed a multi-user \gls{semcom} system called MU-DeepSC for the \gls{vqa} task, where one user sends images and another transmits text-based questions to inquire about the sent images. \glspl{lstm} and \glspl{cnn} are used for textual and visual data encoding, respectively, while their encoders and decoders are jointly optimized to capture the correlation among different data modalities. To improve \cite{xie2021task}, \citet{xie2022task} proposed to unify the architectures and use a transformer-based \dnn regardless of the data modality. However, \citet{xie2022task} presented different \dnn models to perform various tasks which limits the versatility of their method. To address this issue, \citet{zhang2024unified} developed a unified multi-task and multi-modal \gls{semcom} system (U-DeepSC), where a similar end-to-end framework can serve diverse tasks. Moreover, a dynamic scheme is proposed in \cite{zhang2024unified} to adjust the number of transmitted symbols based on the task.

There are also other work \cite{shao2022task} that investigate multi-device cooperative scenarios leveraging \gls{dib} framework \cite{aguerri2019distributed} to achieve distributed feature encoding. \citet{shao2022task} also developed a selective retransmission mechanism to identify the redundancy in the encoded representation of different users which can be further utilized to reduce the communication overhead. On the other hand, \citet{shao2023task} investigated \gls{semcom} systems for temporally-correlated data, where the transmitter removes the video redundancy in both spatial and temporal domains and sends minimal information for the downstream video analysis task. To this end, they proposed a temporal entropy model which uses the features of previous frames as side information to extract the semantics for the current frame. Furthermore, a spatio-temporal fusion module is designed in \cite{shao2023task} to integrate all the features and perform inference at the receiver side.

\subsubsection{\rev{Research Challenges}} \label{sec:challenge3}

\rev{\textbf{Channel Model Inaccuracies.}~All the existing static \gls{semcom} systems use a stochastic model for the channel distortions \cite{bourtsoulatze2019deep, xie2021deep, kurka2020deepjscc}. However, these models are not accurate enough to capture the real-world effects of a dynamic communication environment. Even though dynamic methods try to address this issue by adapting \gls{semcom} systems after deployment \cite{yang2022deep, shao2021learning}, they still initialize the semantic encoder and decoder using an analytical channel model used during training. To address this challenge, recent studies propose adopting an additional data-driven model for the channel \cite{ye2020deep, o2019approximating}. However, this approach adds more computational burden to the developed \gls{semcom} systems, making them difficult to execute and adapt in real-time, especially in resource-constrained devices.}

\rev{\textbf{Task and Modality Dependencies.}~One of the major challenges regarding the proposed \gls{semcom} systems is that they require different \gls{dnn} models for different tasks and data modalities \cite{xie2022task}. Consequently, multiple \glspl{dnn} need to be stored to process various data modalities, or the \gls{dnn} has to be updated when the desired task changes. Therefore, a unified \gls{semcom} system that supports multi-modal data and multiple tasks needs to be studied. To this date, even though several efforts have been made to develop multi-modal and multi-task \gls{semcom} systems \cite{zhang2024unified, jiang2024large}, the desired set of input and output modalities and tasks cannot be dynamically changed at test time.}

\rev{\textbf{Optimization Complexity.}~Existing studies need to learn \gls{semcom} systems in an end-to-end manner by simultaneously sampling over the data and channel distributions. However, the end-to-end design of the \gls{semcom} encoder, precoder, postcoder, and decoder that can be adaptive to both the data set and channel realizations results in huge training overhead as the learning and communication objectives may not be in line with each other \cite{cai2025end}. To address this, \citet{cai2024multi} proposed a decoupled design, in which the learning codecs and communication blocks are trained separately. However, their proposed method suffers from a performance drop as they used a surrogate optimization objective instead of the end-to-end learning objective.}

\begin{table}[!t]
    \centering
    \caption{Summary of Literature on Semantic Communication Systems \label{tab:semcom}}
    \resizebox{\linewidth}{!}{
    \begin{tabular}{ccccc}
    \toprule
    \textbf{Work} & \textbf{Approach} & \textbf{Optimization Objective} & \textbf{Application} & \textbf{Type} \\
    \midrule
    \cite{farsad2018deep} & \makecell{\gls{jscc}, \\ \gls{lstm}} & \gls{wer} & Text Transmission & \makecell{Static, \gls{sl}} \\
    \cite{xie2021deep} & Transformer & \makecell{\gls{bleu}, BERT \\Sentence Similarity} & Text Transmission & \makecell{Static, \gls{sl}} \\
    \cite{xie2020lite} & \makecell{Transformer, \\Pruning, \\Quantization} & \gls{bleu}, MSE & \makecell{Text Transmission \\for \gls{iot}} & \makecell{Static, \gls{sl}} \\
    \cite{sana2022learning} & \makecell{\gls{jscc}, \\Transformer} & \makecell{\gls{bleu}, \\Symbols per Word} & Text Transmission & \makecell{Static, \gls{sl}} \\
    \cite{bourtsoulatze2019deep} & \makecell{\gls{jscc}, \\ \gls{cnn}} & \gls{psnr} & Image Transmission & \makecell{Static, \gls{sl}} \\
    \cite{kurka2020deepjscc} & \gls{jscc}, \gls{cnn} & \gls{psnr}, Bandwidth Ratio & Image Transmission & \makecell{Static, \gls{sl}} \\
    \cite{zhang2022wireless} & \gls{jscc}, \gls{cnn}, \gls{lstm} & \gls{psnr}, \gls{ssim} & Image Transmission & \makecell{Static, \gls{sl}} \\
    \cite{lu2021reinforcement} & \gls{jscc}, \gls{drl} & \gls{wer}, \gls{bleu}, \gls{cider} & Text/Image Transmission & \makecell{Static, \gls{sl}} \\
    \cite{lu2022rethinking} & JSNC, \gls{drl} & \gls{wer}, \gls{bleu} & Text Transmission & \makecell{Static, \gls{sl}} \\
    \rev{\cite{zhao2024lamosc}} & \rev{\gls{llm}, Transformer} & \rev{\gls{psnr}, \gls{ssim}, \gls{lpips}} & \rev{Image Transmission} & \rev{Static, \gls{sl}} \\
    \rev{\cite{jiang2024large}} & \rev{Diffusion, \gls{cgan}} & \rev{\makecell{Accuracy, \\Compression}} & \rev{Multi-Modal Transmission} & \rev{Static, \gls{sl}} \\
    \rev{\cite{zhang2025semantic, wu2024cddm}} & \rev{\makecell{Transformer, \\Diffusion}} & \rev{\makecell{\gls{psnr}, \gls{lpips}, \\MSE, \gls{ssim}}} & \rev{Image Transmission} & \rev{Static, \gls{sl}} \\
    \cite{huang2022toward} & \makecell{\gls{drl}, \gls{gan}, \\Adaptive Quantization} & \gls{psnr}, \gls{ssim}, mAP & \makecell{Image Transmission + \\Segmentation/ \\Object Detection} & \makecell{Static, \gls{sl}/\gls{tl}} \\
    \rev{\cite{bo2024joint}} & \rev{\makecell{Digital Modulation, \\\gls{vae}}} & \rev{\gls{psnr}, Accuracy} & \rev{\makecell{Image Transmission, \\Image Classification}} & \rev{Static, \gls{sl}/\gls{tl}} \\
    \rev{\cite{park2024joint}} & \rev{\makecell{\gls{jscc}, Digital Modulation}} & \rev{\gls{psnr}, Accuracy, mAP} & \rev{\makecell{Image Transmission, \\Image Classification}} & \rev{Static/Adaptive, \gls{sl}/\gls{tl}} \\
    \rev{\cite{tang2024contrastive}} & \rev{\makecell{Contrastive Learning, \\\gls{cnn}}} & \rev{\gls{psnr}, \gls{ssim}, Accuracy} & \rev{\makecell{Image Transmission, \\Image Classification}} & \rev{Static, \gls{sl}/\gls{tl}} \\
    \rev{\cite{zhang2024scan}} & \rev{\makecell{\gls{mimo}, Adaptive Feedback, \\Knowledge Distillation}} & \rev{\gls{psnr}, \gls{csi} Length} & \rev{Image Transmission} & \rev{Adaptive, \gls{sl}} \\
    \cite{zhou2021semantic} & \makecell{\gls{jscc} \\Universal Transformer} & SER, \gls{bleu} & Text Transmission & \makecell{Adaptive, \gls{sl}} \\
    \cite{zhou2022adaptive} & \makecell{\gls{jscc}, Transformer, \\Hybrid \gls{arq}} & Bits per Word, \gls{bleu} & Text Transmission & \makecell{Adaptive, \gls{sl}} \\
    \cite{jiang2022deep} & \makecell{Transformer, \\Hybrid \gls{arq}} & \gls{bleu}, \gls{wer}, Avg Bits & Text Transmission & \makecell{Adaptive, \gls{sl}} \\
    \cite{jankowski2020deep} & \gls{jscc} & Accuracy & Image Retrieval & Static, \gls{tl} \\
    \cite{kang2022task} & \gls{jscc}, \gls{drl} & Accuracy, Latency & Scene Classification for \gls{iot} & \makecell{Static, \gls{tl}} \\
    \cite{shao2021learning} & \gls{jscc}, \gls{ib}, Pruning & Accuracy, Rate & Image Classification & \makecell{Static/Adaptive, \gls{tl}} \\
    \cite{xie2022task} & Transformer & \makecell{Recall, \gls{bleu}, \\Answer Accuracy} & \makecell{\gls{vqa}, Image Retrieval, \\Machine Translation} & \makecell{Static, \gls{tl}} \\
    \cite{zhang2024unified} & Transformer, Codebook & \gls{psnr}, Accuracy, \gls{bleu} & \makecell{\gls{vqa}, Sentiment Analysis, \\Image Classification} & \makecell{Static, \gls{tl}} \\
    \cite{shao2022task} & \gls{cnn}, Attention, \gls{dib} & Accuracy, Rate & \makecell{Multi-view Object Recognition, \\Image Classification} & \makecell{Static, \gls{tl}} \\
    \cite{shao2023task} & \makecell{Entropy Model, \gls{ib}, \\Quantization} & mAP, Latency & Video Analytics & \makecell{Static, \gls{tl}} \\
    \cite{dorner2017deep} & Fine-tuning & \gls{ber}, BLER & Data Transmission & \makecell{Adaptive, Bit-Level} \\
    \cite{aoudia2019model, grathwohl2017backpropagation, aoudia2018end} & Fine-tuning, \gls{drl} & \gls{ber}, BLER & Data Transmission & \makecell{Adaptive, Bit-Level} \\
    \cite{raj2018backpropagating} & Latent Perturbation & \gls{ber}, BLER & Data Transmission & \makecell{Adaptive, Bit-Level} \\
    \cite{o2018physical, o2019approximating} & \makecell{Surrogate Channel \dnn, \\ \gls{gan}} & \gls{ber}, BLER, \gls{gan} Loss & Data Transmission & \makecell{Adaptive, Bit-Level} \\
    \cite{ye2018channel, ye2020deep} & \makecell{\gls{cgan}, \\Pilot Conditioning} & \gls{ber}, BLER, \gls{cgan} Loss & Data Transmission & \makecell{Adaptive, Bit-Level} \\
    \cite{park2020end} & \makecell{Meta-Learning, \\Decoder \gls{tta}} & \gls{ber}, BLER & Data Transmission & \makecell{Adaptive, Bit-Level} \\
    \cite{assine2023slimmable} & Knowledge Distillation & mAP, Latency & Object Detection & \makecell{Adaptive, \gls{tl}} \\
    \cite{abdi2023channel} & \makecell{Dynamic Pruning, \\Ensemble Training} & \makecell{Accuracy, \\Compression, \\Computation} & Image Classification & \makecell{Adaptive, \gls{tl}} \\
    \rev{\cite{cai2024multi, cai2025end}} & \rev{\makecell{\gls{mimo}, Deep Unfolding, \\Surrogate Design Goal}} & \rev{\makecell{Accuracy, Rate, \\Latency}} & \rev{Multi-View Image Classification} & \rev{Adaptive, \gls{tl}} \\
    \bottomrule
    \end{tabular}
    }
\end{table}

\subsection{Advancement in Test-Time Adaptive Systems}

\subsubsection{Channel-Adaptive Coding}

\textbf{Image transmission.}~As the parties in a \gls{semcom} system experience a dynamic wireless link, the entire channel model changes over time which requires the encoder and decoder to be updated during test time accordingly. To address this issue, \citet{dorner2017deep} proposed a two-stage training procedure, where first the entire \dnn is trained using a stochastic channel model which resembles the channel behavior as closely as possible. Then after deployment, only the receiver \dnn is fine-tuned based on the real channel to speed up the adaptation process. Other work \cite{aoudia2019model, grathwohl2017backpropagation, aoudia2018end} utilized reinforcement learning to adapt the encoder \dnn weights to channel variations instead of supervised learning. To this end, \citet{aoudia2018end} regard the transmitter as a RL agent interacting with the environment consisting of both the channel and receiver. Therefore, the agent learns how to update the encoder \dnn parameters to minimize the loss function provided by the environment. Moreover, techniques based on perturbing the latent \cite{raj2018backpropagating} are also proposed to obtain an estimate of the instantaneous channel gradients which are then used to backpropagate and update the transmitter \dnn gradients. \citet{o2018physical} proposed to learn a full-scale surrogate \dnn to accurately model the distribution representing the channel. As such, \glspl{gan} are leveraged to make the surrogate \dnn mimic the ongoing channel behavior \cite{o2019approximating}. Similarly, \citet{ye2018channel} leveraged \glspl{cgan} for time-varying channel modeling where the received signal corresponding to the pilot data is used as the conditioning information \cite{ye2020deep}. Inspired by \dnn \gls{tta} techniques, \citet{park2020end} send multiple pilots over changing channel conditions during the training phase, and utilize the model-agnostic meta-learning (MAML) algorithm \cite{finn2017model} to meta-train an effective adaptation rule for the decoder. As such, the decoder \dnn can be quickly adapted using the learned rules during test time with minimal modifications. \rev{\citet{cai2024multi} proposed a \gls{mimo} \gls{semcom} system by adding a \gls{mimo} precoder block after the semantic encoder, where the precoder takes the \gls{csi} as an input and can be adapted based on the ongoing channel response. As the design objectives of the learning (i.e., semantic encoder and decoder) and communication (i.e., \gls{mimo} precoder) modules are different in previous studies, the authors of \cite{cai2024multi} align such goals by using a unified surrogate measure and separately optimizing these modules. \citet{cai2025end} further improved the alignment of these design goals through a second phase of training using the end-to-end learning objective. In addition, an information-theoretic formulation is provided in \cite{cai2025end} to back up the proposed training method.}

To improve the resilience of \gls{semcom} systems to erratic changes in the wireless environment, even the encoder output size determining the amount of data to transmit needs to be adapted according to varying channel quality. For example, in low \gls{snr} scenarios, the encoded semantics may require more redundancy to achieve a satisfactory task performance while in high \glspl{snr} the semantics can be compressed more to lower the latency. Inspired by this idea, \citet{abdi2023channel} proposed a dynamic pruning approach to adaptively compress the encoder's output dimension according to the ongoing channel conditions. Similarly, \citet{shao2021learning} trained a dynamic activation layer to change the latent length representing the symbols at the channel coding output. \citet{assine2023slimmable} applied knowledge distillation to train the dynamic feature extractor. \citet{hossain2023flexible} proposed to dynamically change the compression rate of entropy coding by controlling the trade-off parameter $\beta$ in Eqn.~\ref{rd_image}. An adaptive rate control mechanism for image transmission is proposed in \cite{yang2022deep}, where a policy \dnn dynamically adjusts the number of active intermediate features based on the channel \gls{snr}.

\rev{Other approaches focus on making the channel feedback size adaptive with respect to the current channel conditions. For example, the authors in \cite{zhang2024scan} proposed a \gls{mimo} \gls{semcom} system, called SCAN, with a lightweight performance evaluator module that is integrated into the transmitter and can predict the reconstruction quality of each image. The compression level of the CSI feedback matrix is then adjusted based on the predicted performance.}

\textbf{Text transmission.}~Unlike some existing methods \cite{xie2021deep, xie2020lite} which use a fixed transformer-based encoder, \citet{zhou2021semantic} adopted a universal transformer to make the encoder adaptive to channel variations and semantic complexity of the input text. More specifically, universal transformer features a circulation mechanism which allows the \dnn structure and its computation time to be adjusted according to the input semantic information and various channel conditions. Similarly, \citet{zhou2022adaptive} proposed an adaptive bit rate control mechanism for the transmission of semantic information, where a policy network is employed to select the optimal rate for the variable-rate encoder.

Retransmission-based techniques are also proposed to improve the reliability of \gls{semcom} frameworks which leads to variable-length encoded semantics based on the number of retransmissions. For example, \citet{jiang2022deep} proposed to combine semantic coding with Reed Solomon channel coding and \gls{arq} which is widely-used in conventional communication systems. More specifically, multiple isomorphic semantic encoders and decoders are used in \cite{jiang2022deep}, where after one decoder fails to interpret the correct semantics and send acknowledgment (ACK), the source message is re-encoded via the next encoder. Moreover, the code length at the output of the transformer-based semantic encoder is made adaptive to the input sentence length.

\subsubsection{Importance-Aware Coding}

As the successful interpretation of some encoded semantics at the receiver side is more important than the others, another line of research focuses on adaptively sending the semantics according to their importance. As such, by prioritizing the transmission of most relevant semantics, performance loss can be mitigated in severe communication environments. For example, \citet{huang2020clio} proposed a progressive slicing algorithm that divides the latent representation into multiple chunks and transmits the semantic slices progressively. \citet{xiao2024nair} demonstrated that the approach in \citep{huang2020clio} is equivalent to a dynamic pruning solution. Therefore, they leveraged the magnitude of \dnn weights to evaluate the relative importance of encoded semantics and dynamically pruned unimportant features in a bandwidth-constrained setting. \citet{hojjat2024limitnet} utilized saliency detection \citep{qin2021boundary} to detect the important part of semantics and gradual ordering \citep{koike2020stochastic} to determine the priority for transmission. \citet{huang2022real} utilized explainable AI tools \citep{davies2021advancing, sundararajan2017axiomatic} to evaluate the importance of latent representations. The most important semantics are processed in the local device while other features are transmitted to the edge. Thus, it can avoid data corruption on important features due to an unreliable wireless link. \citet{wang2021performance} proposed to measure the semantic similarity based on a semantic knowledge graph. After estimating the similarity, an attention-based reinforcement learning algorithm is trained to evaluate the importance and allocate resources for the \gls{semcom} system. \citet{hu2023scalable} investigated the importance-aware \gls{semcom} system in multi-task scenarios. A gradient-based importance evaluation tool \citep{selvaraju2017grad} is leveraged to evaluate the importance of multi-task semantics after the \gls{jscc} block to achieve a scalable channel coding.

Modern wireless communication systems often leverage \gls{ofdm} and \gls{mimo} techniques. In order to incorporate \gls{semcom} into advanced wireless technologies, several importance-aware semantic transmission approaches are proposed. \citet{zhou2024feature} proposed a channel prediction algorithm to estimate the future \gls{csi} in \gls{mimo} systems. Then, the sub-carriers having better predicted channel conditions are allocated to the important semantics. \citet{weng2024semantic} investigated the importance-aware \gls{semcom} for both single-user and multi-user \gls{mimo}. A neural network-based channel estimation algorithm and a singular value decomposition precoding are used to jointly determine the sub-carrier allocation for separate semantics. \citet{liu2024ofdm} studied \gls{ofdm} systems and leveraged \gls{drl} to jointly determine the sub-carrier allocation and adaptive quantization based on the semantic importance.

The challenge faced when developing test-time adaptive \gls{semcom} systems is the extra cost of tracking the channel variations or evaluating the importance of encoded semantics to dynamically update the encoder and decoder \dnns. This extra computation burden may become a critical bottleneck when deploying such techniques on resource-constrained devices. On the other hand, while static approaches do not pose such challenges, their performance degrades noticeably after deployment.

\subsubsection{\rev{Research Challenges}} \label{sec:challenge4}

\rev{\textbf{Adaptation Complexity.}~Even though dynamic \gls{semcom} systems are developed to tackle distribution shifts in data and communication environments \cite{cai2025end, park2020end}, the proposed adaptation mechanisms are computationally expensive. This leads to an intolerable latency in the proposed \gls{semcom} systems. Also, such systems fail to respond quickly after a major test-time change. Recent studies~\cite{zhang2024o2sc} address this challenge by introducing a separate lightweight adaptive module instead of adapting the entire \gls{semcom} transmitter and receiver chain.}

\rev{\textbf{Changing Objectives.}~The existing works focus on adapting the \gls{semcom} systems only based on the observations from the communication channel and input data \citet{cai2024multi, o2019approximating}. However, user requirements, communication constraints, resource availability, and device workload may change during test time. Therefore, \gls{semcom} systems should be also adaptable based on high-level commands or waveform objectives.}

\rev{\textbf{Security Threats.}~\Gls{semcom} systems introduce unique security challenges due to their different design goals compared to traditional communication systems. For example, \gls{semcom} systems with adaptive rate control mechanisms are proposed, where the number of transmitted symbols depends on the difficulty of the message \citet{yang2022deep, shao2021learning}. As such, an adversary can focus on injecting semantic noise at the transmitter to compromise the communication efficiency of \gls{semcom} systems, apart from their performance. In addition, the attacker's objective can be the maximization of the semantic content of the stolen data rather than its quantity. Moreover, stealing the knowledge bases in the \gls{semcom} systems that employ them can also be identified as a potential threat. On the other hand, \gls{semcom} systems are vulnerable to over-the-air attacks (attacks in the latent space) which damage their performance \cite{zhang2023adversarial}. Another topic that requires further investigation is rendering adversarial transmitters and receivers unable to communicate with the \gls{semcom} parties, where an encryption scheme could be potentially employed to establish a secure \gls{semcom} link.}

\rev{\textbf{Lack of Higher-Level Networking Functionalities.}~Studied \gls{semcom} systems assume a direct link between the transmitter and receiver \citet{abdi2025phydnns}. However, retransmission and routing mechanisms need to be developed so that nodes that are multiple hops apart in a \gls{semcom} network can communicate with each other. Moreover, multi-user semantic signal fusion/defusion is another challenging area that should be explored more in depth.}

\section{Future Research Directions} \label{sec:discuss}

\subsection{Co-Design of Compression, Partitioning, and Communication}

\rev{Recent studies have achieved significant advances in compression, partitioning, and communication for distributed \dnns. However, most existing research focuses on optimizing only one of these areas in isolation, neglecting potential synergies between them. This narrow focus is understandable given the inherent complexity of joint optimization, which demands extensive cross-domain expertise spanning both \gls{sec} and \gls{semcom}, along with comprehensive system design.}

\rev{As outlined in Section~\ref{sec:unify}, \gls{sec} and \gls{semcom} can substantially benefit from integrated designs that unify optimization approaches across both domains. While limited, some research has successfully demonstrated joint optimization can achieve superior performance compared to single-objective approaches. For instance, \citet{mu2023fine} proposed a novel optimization framework to jointly compress the semantic and optimize the \dnn partitioning, which achieving smaller latency compared to prior studies that solely focuses on partitioning \cite{kang2017neurosurgeon} or compression \cite{shao2020bottlenet++}. Also, \citet{abdi2023channel} explored feature compression and channel-adaptive coding, proposing an adaptive compression method based on pruning that effectively enhances \dnn inference latency in dynamic wireless environments. Therefore, a promising research direction could involve the co-design of compression, partitioning, and communication.}

\subsection{Security and Privacy of Semantic Systems}

\rev{While distributing \dnns can achieve real-time performance in resource-constrained environments, the distributed nature also expose the semantic information to adversaries, raising security concerns in these semantic systems. An adversary can eavesdrop the semantics and add an adversarial perturbation to compromise the task-oriented performance. A prior study \cite{zhang2023adversarial} provided a rigorous theoretical analysis for adversarial robustness of latent representations using \gls{ib}. In addition, it provided a large-scale empirical study to assess the robustness of multiple semantic compression approaches. However, \cite{zhang2023adversarial} leveraged existing attack algorithms in conventional adversarial machine learning domain which may not be optimized for attacks on the semantics. The adversarial robustness of \gls{sec} and \gls{semcom} need a more comprehensive assessment with more advanced attack methods.}

\rev{In addition, adversarial attacks can compromise not only the end-to-end task performance but also the in-network performance. For example, \citet{zhang2024resilience} revealed that entropy coding methods generating variable-length semantics are vulnerable to both noise and adversarial perturbations in the input space. Specifically, adversaries can degrade network performance by manipulating inputs to increase the entropy of latent representations, thereby exploiting the variable-length nature of the coding scheme.}

\rev{Moreover, the cooperative manner of \dnn inference also creates additional privacy risks. Adversaries can potentially reconstruct original input data from exposed semantic information \cite{he2019model,he2020attacking,chen2024dia}. This reconstruction capability poses serious privacy concerns, particularly in applications involving sensitive personal data.}

\rev{Current research in \gls{sec} and \gls{semcom} has predominantly focused on optimizing computational and communication efficiency while largely overlooking these critical security dimensions. We argue that integrating robust learning techniques and privacy-preserving mechanisms represents a compelling and necessary research direction for advancing secure collaborative intelligence systems.}

\subsection{Multi-Device Collaborative Inference}

\rev{Most of the existing literature in \gls{sec} and \gls{semcom} only consider the scenario where \dnns are deployed on a single mobile and a single edge device. This assumes the mobile-edge system has enough resource to execute large \dnn models. However, this assumption becomes increasingly unrealistic, as individual mobile devices and edge servers often lack the computational capacity required for modern large-scale models, particularly contemporary large language models and vision foundation models that demand extensive computational resources \cite{dubey2024llama,brown2020language}. As such, it is worth to extend the distributed inference to a more realistic case where multiple devices with heterogeneous resources are used to perform the same task collaboratively.}

\rev{Some literature has investigated multi-device collaborative inference. For instance, \citet{mohammed2020distributed} proposed to divide \dnns into multiple parts and dynamically deploy them using a matching algorithm. \citet{disabato2021distributed} implemented the multi-device partitioning on a real \gls{iot} system. \citet{hu2022distributed} considered to split \dnn into multiple parts based on its \gls{dag} structure. The optimization problem is formulated as a linear programming problem and an off-the-shelf solver is leveraged to find the solution. \citet{jung2023optimization} considered the distributed \dnn inference as a routing problem and proposed a layered graph model to optimize the data transfer and computing node selection. \citet{li2024distributed} proposed to use \gls{drl} to achieve a more fine-grained partitioning for complex \dnn architectures where conventional approaches may need an exhaustive search.}

\rev{While these studies demonstrated promising results, they face substantial barriers to real-world deployment. The increasing number of devices and heterogeneous operating conditions make the optimization problem more challenging compared to the two-device partitioning scenario. This becomes more severe when the \dnn has a large size. Existing work only demonstrated the effectiveness of their methods through simulation or simple \dnn architectures \cite{mohammed2020distributed,disabato2021distributed}. Moreover, the increased problem complexity requires longer optimization times, making it difficult to quickly adapt to changing wireless environments. \cite{hu2022distributed,jung2023optimization,li2024distributed}. This research gap highlights the necessity for more efficient and practical solutions for multi-device collaborative inference.}



\subsection{Multi-Task Scalability/Multi-Modal Versatility}

\rev{Existing research in both \gls{sec} and \gls{semcom} tend to develop systems optimized for a single task. For example, \citet{matsubara2022bottlefit} optimized the \dnn architecture and training method xclusively for image classification tasks, while the \gls{semcom} framework proposed in \cite{shao2021learning} employed distinct \gls{dnn} models for different datasets. These separated systems will creates substantial computational overhead when multiple tasks must be executed simultaneously as they require to run multiple \glspl{dnn} separately. This presents a critical barrier in deployment on resource-constrained devices. As such, it urges future research to investigate more general frameworks that are optimized for multiple tasks simultaneously and can effectively handle multi-modal data inputs. }

\rev{While limited, some recent work has successfully demonstrated the feasibility of multi-task and multi-modal learning in \gls{sec} and \gls{semcom}. For instance, \citet{matsubara2025multi} demonstrated that the proposed multi-task semantic compression can effectively reduce end-to-end latency by up to 95\% and energy consumption by up to 88\% in a joint image classification, object detection, and semantic segmentation scenario. \citet{zhang2024unified} proposed to use a single \dnn with dynamic feature selection to handle multi-modal and multi-task applications, achieving comparable performance to task-specific models while using 71\% fewer computation. These promising results underscore the potential of unified approaches to address the scalability challenges inherent in current task-specific semantic systems.}

\subsection{Real-World Applications}

\rev{As discussed in Section~\ref{sec:background}, \gls{sec} and \gls{semcom} play significant roles in achieving edge intelligence in 6G networks. However, current research has predominantly validated their frameworks using computer vision and natural language processing datasets, creating a notable gap between the demonstrated capabilities and the diverse requirements of real-world applications.}

\rev{Several recent studies have begun exploring more specialized applications. For example, \citet{bahadori2023splitbeam} used semantic compression to compress the Wi-Fi beamforming feedback information which achieves up to 80\% transmission reduction compared to the conventional matrix decomposition approaches. \citet{liu2023accelerating} investigated the model partitioning in vehicular networks to achieve low latency and high reliability in Vehicle-to-Vehicle/Infrastructure communication. \citet{li2024optimal} optimized the \dnn splitting point and resource allocation to minimize the energy consumption of sensors in multi-device \gls{csi} sensing. }

\rev{These domain-specific implementations demonstrate the broader applicability of \gls{sec} and \gls{semcom} beyond traditional computer vision and NLP tasks. Nevertheless, numerous other potential applications remain largely unexplored such as healthcare monitoring, smart home systems, and industrial IoT, all of which could significantly benefit from \gls{sec} and \gls{semcom} frameworks. Future research should systematically investigate these diverse application domains to fully realize the potential of semantic approaches in 6G networks.}

\section{Conclusion} \label{sec:conclude}

\acrlong{sec} and \acrlong{semcom} are two critical approaches to achieve collaborative intelligence for next-generation edge computing. However, there is no existing survey to systematically unify both fields and literature in one domain often neglects the progress in the other. In this work, we provide a unified overview and review the recent breakthroughs in both fields by addressing their technical contributions. In addition, we discuss the research challenges and potential directions for future endeavors. We hope this survey can inspire further research for collaborative intelligence. 

\section*{Acknowledgements}

This work was supported in part by the U.S. Army Combat Capabilities Development Command (DEVCOM) Army Research Laboratory under Cooperative Sub-Award Agreement No. W911NF-24-2-0224, by the National Science Foundation under grants CNS-2134973, ECCS-2229472, by the Air Force Office of Scientific Research under contract number FA9550-23-1-0261 and in part by the Office of Naval Research under award number N00014-23-1-2221.

\bibliographystyle{elsarticle-num-names.bst}
\bibliography{bibliography}

\end{document}